\documentclass[runningheads]{llncs}
\usepackage{graphicx}
\usepackage{tikz}
\usepackage{comment} 
\usepackage{amsmath,amssymb} % define this before the line numbering.
\usepackage{color}

\usepackage[linesnumbered,boxed]{algorithm2e}
\SetKwInput{KwInput}{Input}
\SetKwInput{KwOutput}{Output}
\usepackage[export]{adjustbox}
\usepackage{subfig}
\usepackage[inline]{enumitem}
\usepackage{booktabs}
\usepackage{array,multirow,graphicx}

\usepackage[implicit=false]{hyperref}
\hypersetup{
    colorlinks=true,
    urlcolor=blue,
}

\RequirePackage{color}
\definecolor{RED}{rgb}{1,0,0}
\definecolor{BLUE}{rgb}{0,0,1}
\definecolor{White}{rgb}{1,1,1}

\def\eg{\emph{e.g.,}} 
\def\ie{\emph{i.e.,}}

\def\vs{\emph{vs. }}

\def\etal{\emph{et al.}}

\def\Vec#1{{\boldsymbol{#1}}}
\def\Mat#1{{\boldsymbol{#1}}}

\newcommand{\fig}{Fig.}
\newcommand{\tab}{Table}

\usepackage[font=small,labelfont=bf,tableposition=top]{caption}

\begin{document}
% \linenumbers
\pagestyle{headings}
\mainmatter
\def\ECCVSubNumber{445}  % Insert your submission number here

\title{AiR: Attention with Reasoning Capability} % Replace with your title

\begin{comment}
\titlerunning{ECCV-20 submission ID \ECCVSubNumber} 
\authorrunning{ECCV-20 submission ID \ECCVSubNumber} 
\author{Anonymous ECCV submission}
\institute{Paper ID \ECCVSubNumber}
\end{comment}
%******************

%\begin{comment}
\titlerunning{AiR: Attention with Reasoning Capability}
\author{Shi Chen\thanks{Equal contributions.}\orcidID{0000-0002-3749-4767} \and
Ming Jiang\inst{\star}\orcidID{0000-0001-6439-5476} \and
Jinhui Yang\orcidID{0000-0001-8322-1121} \and
Qi Zhao\orcidID{0000-0003-3054-8934}}
\authorrunning{S. Chen et al.}
\institute{
University of Minnesota, Minneapolis MN 55455, USA\\
\email{\{chen4595,mjiang,yang7004,qzhao\}@umn.edu}}
%\end{comment}
%******************
\maketitle

\begin{abstract}
While attention has been an increasingly popular component in deep neural networks to both interpret and boost performance of models, little work has examined how attention progresses to accomplish a task and whether it is reasonable. In this work, we propose an Attention with Reasoning capability (AiR) framework that uses attention to understand and improve the process leading to task outcomes. We first define an evaluation metric based on a sequence of atomic reasoning operations, enabling quantitative measurement of attention that considers the reasoning process. We then collect human eye-tracking and answer correctness data, and analyze various machine and human attentions on their reasoning capability and how they impact task performance. Furthermore, we propose a supervision method to jointly and progressively optimize attention, reasoning, and task performance so that models learn to look at regions of interests by following a reasoning process. We demonstrate the effectiveness of the proposed framework in analyzing and modeling attention with better reasoning capability and task performance. The code and data are available at \url{https://github.com/szzexpoi/AiR}.

\end{abstract}

\section{Introduction}
Recent progress in deep neural networks (DNNs) has resulted in models with significant performance gains in many tasks. Attention, as an information selection mechanism, has been widely used in various DNN models, to improve their ability of localizing important parts of the inputs, as well as task performances. It also enables fine-grained analysis and understanding of the black-box DNN models, by highlighting important information in their decision-making. Recent studies explored different machine attentions and showed varied degrees of agreement on where human consider important in various vision tasks, such as captioning~\cite{captioning_human_19,captioning_human_17} and visual question answering (VQA)~\cite{vqahat}.

\begin{figure}
\centering
\includegraphics[width=1\linewidth]{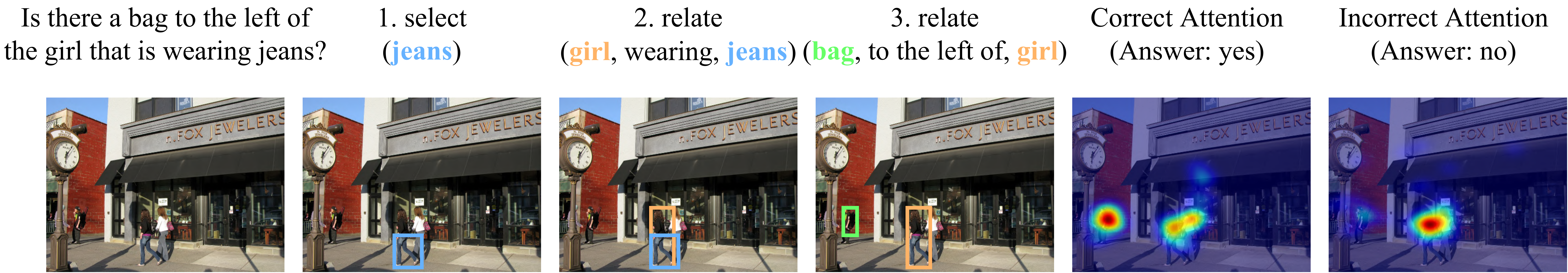}
\caption{Attention is an essential mechanism that affects task performances in visual question answering. Eye fixation maps of humans suggest that people who answer correctly look at the most relevant ROIs in the reasoning process (\ie~jeans, girl, and bag), while incorrect answers are caused by misdirected attention}
\label{fig:teaser}
\end{figure}

Similar to humans who look and reason actively and iteratively to perform a visual task, attention and reasoning are two intertwined mechanisms underlying the decision-making process. As shown in \fig~\ref{fig:teaser}, answering the question requires humans or machines to make a sequence of decisions based on the relevant regions of interest (ROIs) (\ie~to sequentially look for the jeans, the girl wearing the jeans, and the bag to the left of the girl). Guiding attention to explicitly look for these objects following the reasoning process has the potential to improve both interpretability and performance of a computer vision model. 

To understand the roles of visual attention in the visual reasoning context, and leverage it for model development, we propose an integrated Attention with Reasoning capability (AiR) framework. It represents the visual reasoning process as a sequence of atomic operations each with specific ROIs, defines a metric and proposes a supervision method that enables the quantitative evaluation and guidance of attentions based on the intermediate steps of the visual reasoning process. A new eye-tracking dataset is collected to support the understanding of human visual attention during the visual reasoning process, and is also used as a baseline for studying machine attention. This framework is a useful toolkit for research in visual attention and its interaction with visual reasoning.

Our work has three distinctions from previous attention evaluation~\cite{vqahat,gqa,explanation,qual_eval} and supervision~\cite{paan,han,mining} methods: (1) We go beyond the existing evaluation methods that are either qualitative or focused only on the alignment with outputs, and propose a measure that encodes the progressive attention and reasoning defined by a set of atomic operations. (2) We emphasize the tight correlation between attention, reasoning, and task performance, conducting fine-grained analyses of the proposed method with various types of attention, and incorporating attention with the reasoning process to enhance model interpretability and performance. (3) Our new dataset with human eye movements and answer correctness enables more accurate evaluation and diagnosis of attention.

To summarize, the proposed framework makes the following contributions:
\begin{enumerate}
%    \item  It includes
    \item A new quantitative evaluation metric (AiR-E) to measure attention in the reasoning context, based on a set of constructed atomic reasoning operations.
    \item A supervision method (AiR-M) to progressively optimize attention throughout the entire reasoning process.
    \item An eye-tracking dataset (AiR-D) featuring high-quality attention and reasoning labels as well as ground truth answer correctness.
    \item Extensive analyses of various machine and human attention with respect to reasoning capability and task performance. Multiple factors of machine attention have been examined and discussed. Experiments show the importance of progressive supervision on both attention and task performance.
\end{enumerate}

\section{Related Works}\label{sec:related}

This paper is most closely related to prior studies on the evaluation of attention in visual question answering (VQA)~\cite{vqahat,gqa,explanation,qual_eval}. In particular, the pioneering work by Das~\etal~\cite{vqahat} is the only one that collected human attention data on a VQA dataset and compared them with machine attention, showing considerable discrepancies in the attention maps. Our proposed study highlights several distinctions from related works:
\begin{enumerate*}
    \item[(1)] Instead of only considering one-step attention and its alignment with a single ground-truth map, we propose to integrate attention with progressive reasoning that involves a sequence of operations each related to different objects.
    \item[(2)] While most VQA studies assume human answers to be accurate, it is not always the case~\cite{yang2018visual}. We collect ground truth correctness labels to examine the effects of attention and reasoning on task performance.
    \item[(3)] The only available dataset~\cite{vqahat}, with post-hoc attention annotation collected on blurry images using a ``bubble-like'' paradigm and crowdsourcing, may not accurately reflect the actual attention of the task performers~\cite{tracking_compare}. Our work addresses these limitations by using on-site eye tracking data and QA annotations collected from the same participants.% [QZ: can we add one sentence to summarize how we address post-hoc annotation? anything else to add for dataset difference?]. %We collect the first eye-tracking dataset to accurately capture the attention of people answering the same types of questions as VQA models.
    \item[(4)] Das~\etal~\cite{vqahat} compared only spatial attention with human attention. Since recent studies~\cite{gqa,qual_eval} suggest that attention based on object proposals are more semantically meaningful, we conduct the first quantitative and principled evaluation of object-based attentions.
\end{enumerate*}

This paper also presents a progressive supervision approach for attention, which is related to the recent efforts on improving attention accuracy with explicit supervision. Several studies use different sources of attention ground truth, such as human attention~\cite{han}, adversarial learning~\cite{paan} and objects mined from textual descriptions~\cite{mining}, to explicitly supervise the learning of attentions. Similar to the evaluation studies introduced above, these attention supervision studies only consider attention as a single-output mechanism and ignores the progressive nature of the attention process or whether it is reasonable or not. As a result, they fall short of acquiring sufficient information from intermediate steps. Our work addresses these challenges with joint prediction of the reasoning operations and the desired attentions along the entire decision-making process. %It is motivated by the recent successes in decomposing reasoning into different sub-problems~\cite{nmn,n2nmn,tbd-net}. While these works focus on developing compositional reasoning models, our aim is to study the accuracy and effectiveness of different attention mechanisms, based on the decomposed reasoning process.

Our work is also related to a collection of datasets for eye-tracking and visual reasoning. Eye tracking data is collected to study passive exploration~\cite{alers2010studying,borji2015cat2000,fan2018emotional,judd2009learning,koehler2014saliency,xu2014predicting} as well as task-guided attention~\cite{alers2010studying,ehinger2009modelling,koehler2014saliency}. Despite the less accurate and post-hoc mouse-clicking approximation~\cite{vqahat}, there has been no eye-tracking data recorded from human participants performing the VQA tasks. To facilitate the analysis of human attention in VQA tasks, we construct the first dataset of eye-tracking data collected from humans performing the VQA tasks.
A number of visual reasoning datasets~\cite{vqa1,vqa2,gqa,clevr,movieqa,vcr} are collected in the form of VQA. Some are annotated with human-generated questions and answers \cite{vqa1,movieqa}, while others are developed with synthetic scenes and rule-based templates to remove the subjectiveness of human answers and language biases~\cite{vqa2,gqa,clevr,vcr}. The one most closely related to this work is GQA~\cite{gqa}, which offers naturalistic images annotated with scene graphs and synthetic question-answer pairs. With balanced questions and answers, it reduces the language bias without compromising generality. Their data efforts benefit the development of various visual reasoning models~\cite{updown,trilinear,mcb,n2nmn,ban,tbd-net,sc_vqa,san,ns_vqa,coatt,mfb,hint}. In this work, we use a selection of GQA data and annotations in the development of the proposed framework.

\section{Method}\label{sec:framework}
Real-life vision tasks require looking and reasoning interactively. This section presents a principled framework to study attention in the reasoning context. It consists of three novel components:
\begin{enumerate*}
  \item[(1)] a quantitative measure to evaluate attention accuracy in the reasoning context,
  \item[(2)] a progressive supervision method for models to learn where to look throughout the reasoning process, and
  \item[(3)] an eye-tracking dataset featuring human eye-tracking and answer correctness data.
\end{enumerate*}

\subsection{Attention with Reasoning Capability} \label{reasoning_progress}

To model attention as a process and examine its reasoning capability, we describe reasoning as a sequence of atomic operations. Following the sequence, an intelligent agent progressively attends to the key ROIs at each step and reasons what to do next until eventually making a final decision. A successful decision-making relies on accurate attention for various reasoning operations, so that the most important information is not filtered out but passed along to the final step.
\begin{table}[b]
\begin{center}
\caption{Semantic operations of the reasoning process}
\label{semantic}
\resizebox{1\textwidth}{!}{
\begin{tabular}{|c|c|}
\hline
\textbf{Operation} & \textbf{Semantic}  \\
\hline
Select & Searching for objects from a specific category. \\
\hline
Filter & Determining the targeted objects by looking for a specific attribute. \\
\hline
Query & Retrieving the value of a specific attribute from the ROIs. \\
\hline
Verify & Examining the targeted objects and checking if they have a given attribute. \\
\hline 
Compare & Comparing the values of an attribute between multiple objects. \\
\hline
Relate & Connecting different objects through their relationships. \\
\hline
And/Or & Serving as basic logical operations that combine the results of the previous operation(s). \\
\hline
\end{tabular}
}
\end{center}
\end{table}

To represent the reasoning process and obtain the corresponding ROIs, we define a vocabulary of atomic operations emphasizing the role of attention. These operations are grounded on the 127 types of operations of GQA \cite{gqa} that completely represent all the questions. %Each operation is denoted as a universal triplet $<$operation, attribute, category$>$. 
As described in \tab~\ref{semantic}, some operations require attention to a specific object (\textit{query}, \textit{verify}); some require attention to objects of the same category (\textit{select}), attribute (\textit{filter}), or relationship (\textit{relate}); and others require attention to any (\textit{or}) or all (\textit{and}, \textit{compare}) ROIs from the previous operations. The ROIs of each operation are jointly determined by the type of operation and the scene information (\ie~object categories, attributes and relationships). Given the operation sequence and annotated scene information, we can traverse the reasoning process, starting with all objects in the scene, and sequentially apply the operations to obtain the ROIs at each step. Details of this method are described in the supplementary materials.
\begin{figure}[t]
\centering
\includegraphics[width=0.75\linewidth]{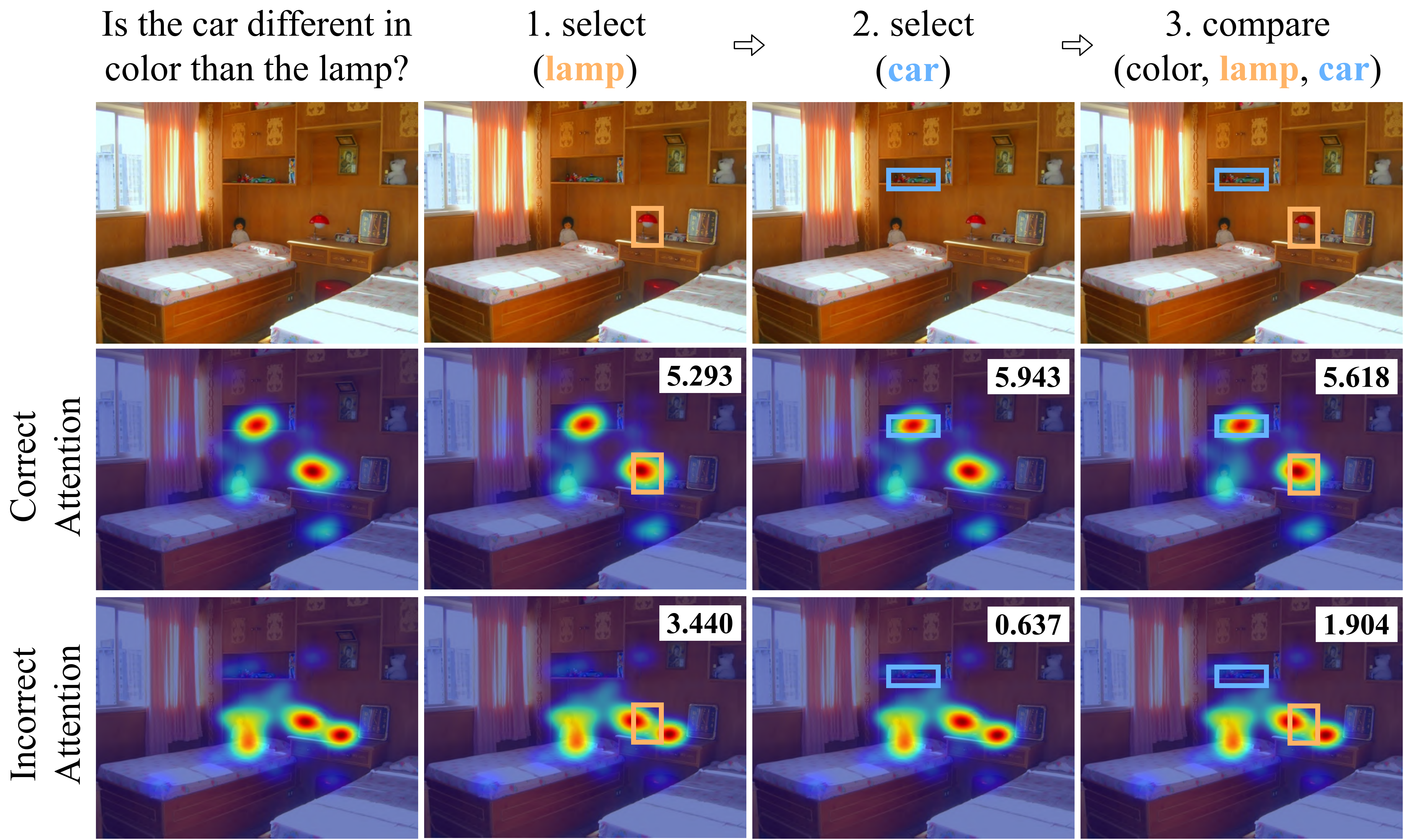}
\caption{AiR-E scores of Correct and Incorrect human attention maps, measuring their alignments with the bounding boxes of the ROIs}
\label{fig:framework}
\end{figure}

\subsection{Measuring Attention Accuracy with ROIs}\label{sec:AiR-E}
Decomposing the reasoning process into a sequence of operations allows us to evaluate the quality of attention (machine and human attentions) according to its alignment with the ROIs at each operation. Attention can be represented as a 2D probability map where values indicate the importance of the corresponding input pixels. To quantitatively evaluate attention accuracy in the reasoning context, we propose the AiR-E metric that measures the alignment of the attention maps with ROIs relevant to reasoning. As shown in \fig~\ref{fig:framework}, for humans, a better attention map leading to the correct answer has higher AiR-E scores, while the incorrect attention with lower scores fails to focus on the most important object (\ie~car). It suggests potential correlation between the AiR-E and the task performance. The specific definition of AiR-E is introduced as follows:

Inspired by the Normalized Scanpath Saliency \cite{sal_metric} (NSS), given an attention map $A(x)$ where each value represents the importance of a pixel $x$, we first standardize the attention map into $A^*(x) = \left(A(x) - \mu\right)/\sigma$,
where $\mu$ and $\sigma$ are the mean and standard deviation of the attention values in $A(x)$, respectively. For each ROI, we compute AiR-E as the average of $A^*(x)$ inside its bounding box $B$: $\text{AiR-E}(B) =  \sum\limits_{x \in B} A^*(x)/|B|$. Finally, we aggregate the AiR-E of all ROIs for each reasoning step:
\begin{enumerate}
    \item For operations with one set of ROIs (\ie~\textit{select}, \textit{query}, \textit{verify}, and \textit{filter}), as well as \textit{or} that requires attention to one of multiple sets of ROIs, an accurate attention map should align well with at least one ROI. Therefore, the aggregated AiR-E score is the maximum AiR-E of all ROIs.
    \item For those with multiple sets of ROIs (\ie~\textit{relate}, \textit{compare}, \textit{and}), we compute the aggregated AiR-E for each set, and take the mean across all sets.
\end{enumerate}

% By considering the different attention requirements of the operations, AiR-E is able to measure the accuracy of the alignment between an attention map and various ROIs at each step of the reasoning process. 

\subsection{Reasoning-aware Attention Supervision} \label{sup_method}
For models to learn where to look along the reasoning process, we propose a reasoning-aware attention supervision method (AiR-M) to guide models to progressively look at relevant places following each reasoning operation. Different from previous attention supervision methods~\cite{paan,han,mining}, the AiR-M method considers the attention throughout the reasoning process and jointly supervises the prediction of reasoning operations and ROIs across the sequence of multiple reasoning steps. Integrating attention with reasoning allows models to accurately capture ROIs along the entire reasoning process for deriving the correct answers.

The proposed method has two major distinctions:
\begin{enumerate*}
    \item[(1)] integrating attention progressively throughout the entire reasoning process and 
    \item[(2)] joint supervision on attention, reasoning operations and answer correctness.
\end{enumerate*} 
Specifically, following the reasoning decomposition discussed in Section~\ref{reasoning_progress}, at the $t$-th reasoning step, the proposed method predicts the reasoning operation $\Vec{r}_{t}$, and generates an attention map $\Vec{\alpha}_{t}$ to predict the ROIs. With the joint prediction, models learn desirable attentions for capturing the ROIs throughout the reasoning process and deriving the answer. The predicted operations and the attentions are supervised together with the prediction of answers:
\begin{equation}
    L = L_{ans} + \theta \sum\limits_{t} L_{\Vec{\alpha_t}} +  \phi \sum\limits_{t}  L_{\Vec{r}_t}
\end{equation}
\noindent where $\theta$ and $\phi$ are hyperparameters. We use the standard cross-entropy loss $L_{ans}$ and $L_{\Vec{r}_t}$ to supervise the answer and operation prediction, and a Kullback–Leibler divergence loss $L_{\Vec{\alpha}_t}$ to supervise the attention prediction. We aggregate the loss for operation and attention predictions over all reasoning steps.

The proposed AiR-M supervision method is general, and can be applied to various models with attention mechanisms. In the supplementary materials, we illustrate the implementation details for integrating AiR-M with different state-of-the-art models used in our experiments.

\subsection{Evaluation Benchmark and Human Attention Baseline}

Previous attention data collected under passive image viewing~\cite{mit_sal}, approximations with post-hoc mouse clicks~\cite{vqahat}, or visually grounded answers~\cite{explanation} may not accurately or completely reflect human attention in the reasoning process. They also do not explicitly verify the correctness of human answers. To demonstrate the effectiveness of the proposed evaluation metric and supervision method, and to provide a benchmark for attention evaluation, we construct the first eye-tracking dataset for VQA. It, for the first time, enables the step-by-step comparison of how humans and machines allocate attention during visual reasoning.

Specifically, we 
\begin{enumerate*}
\item[(1)] select images and questions that require humans to actively look and reason;
\item[(2)] remove ambiguous or ill-formed questions and verify the ground truth answer to be correct and unique;
\item[(3)] collect eye-tracking data and answers from the same human participants, and evaluate their correctness with the ground-truth answers.
\end{enumerate*}

% Our AiR-D dataset accurately captures what humans consider important along the reasoning process, allowing to comparatively analyze both the attention accuracy and the task performance of models and humans, enriching the understanding of visual attention in the reasoning process. 

\textbf{Images and questions.} Our images and questions are selected from the balanced validation set of GQA~\cite{gqa}. Since the questions of the GQA dataset are automatically generated from a number of templates based on scene graphs~\cite{krishna2017visual}, the quality of these automatically generated questions may not be sufficiently high. Some questions may be too trivial or too ambiguous. Therefore, we perform automated and manual screenings to control the quality of the questions. First, to avoid trivial questions, all images and questions are first screened with these criteria:
\begin{enumerate*}
\item[(1)] image resolution is at least 320$\times$320 pixels;
\item[(2)] image scene graph consists of at least 16 relationships;
\item[(3)] total area of question-related objects does not exceed 4\% of the image.
\end{enumerate*}
Next, one of the authors manually selects 987 images and 1,422 questions to ensure that the ground-truth answers are accurate and unique. The selected questions are non-trivial and free of ambiguity, which require paying close attention to the scene and actively searching for the answer.

\textbf{Eye-tracking experiment.} The eye-tracking data are collected from 20 paid participants, including 16 males and 4 females from age 18 to 38. They are asked to wear a Vive~Pro~Eye headset with an integrated eye-tracker to answer questions from images presented in a customized Unity interface. The questions are randomly grouped into 18 blocks, each shown in a 20-minute session. The eye-tracker is calibrated at the beginning of each session. During each trial, a question is first presented, and the participant is given unlimited time to read and understand it. The participant presses a controller button to start viewing the image. The image is presented in the center for 3 seconds. The image is scaled such that both the height and width occupy 30 degrees of visual angle (DVA). After that, the question is shown again and the participant is instructed to provide an answer. The answer is then recorded by the experimenter. The participant presses another button to proceed to the next trial.

\begin{figure}[t]
    \centering
    \subfloat[]{\includegraphics[width=0.3\linewidth]{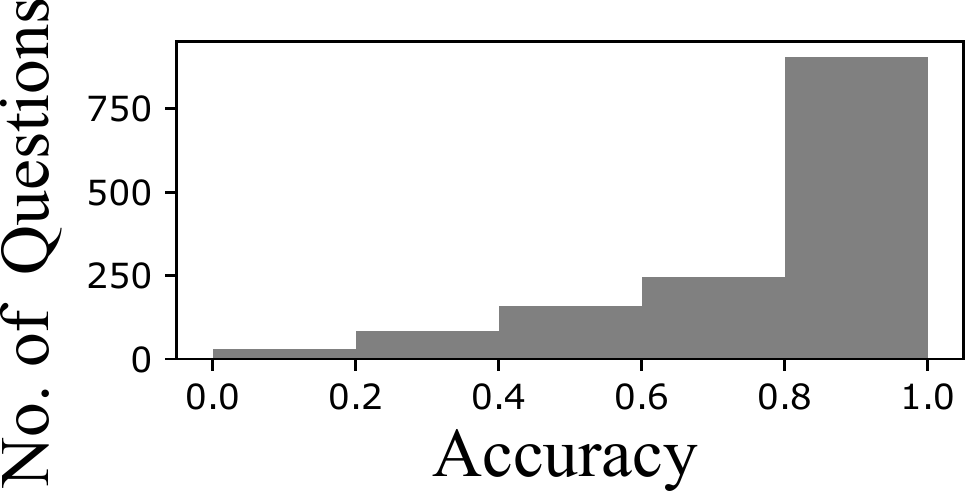}}
    \qquad
    \subfloat[]{\includegraphics[width=0.3\linewidth]{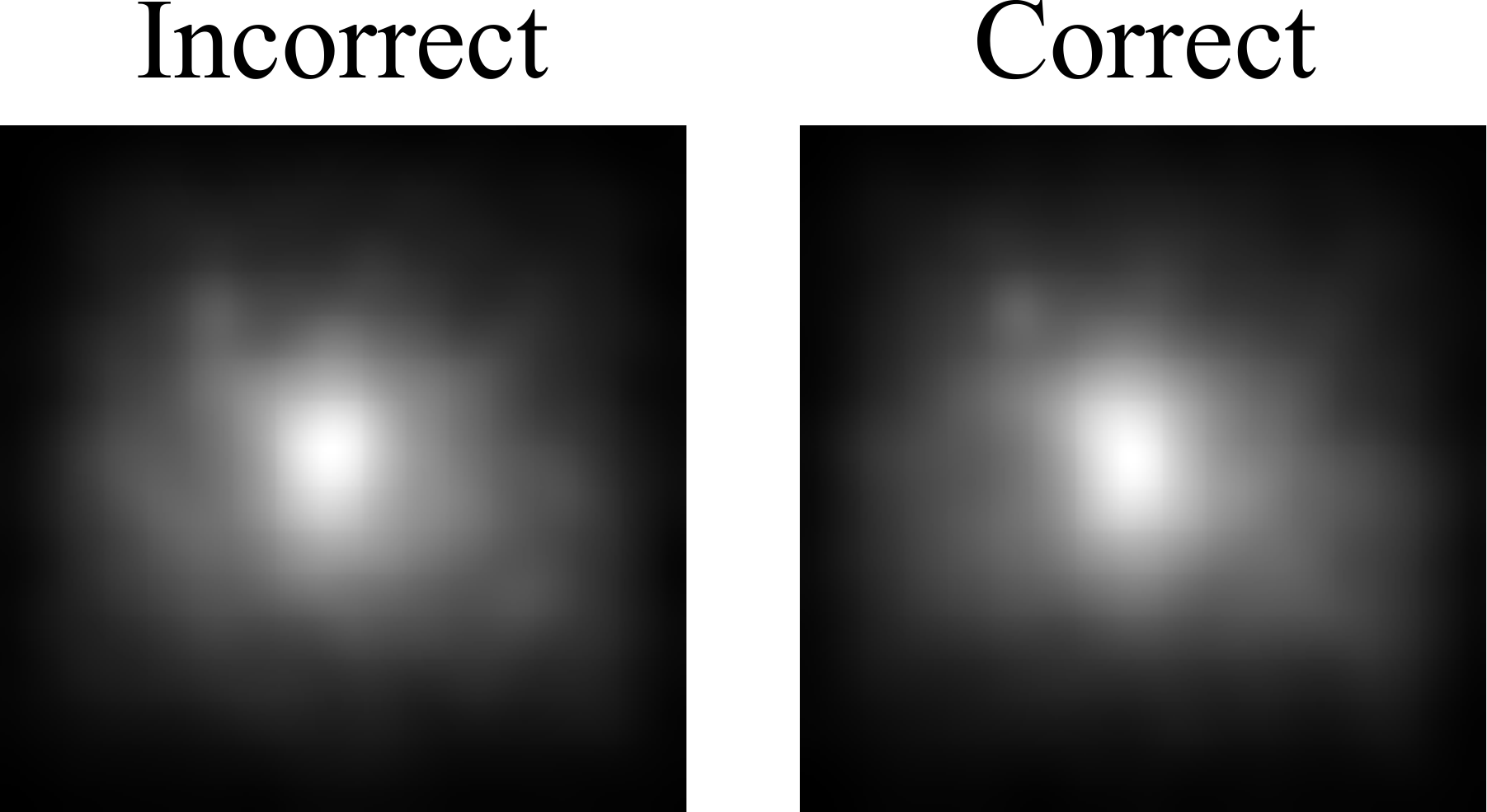}}
    \caption{Distributions of answer accuracy and eye fixations of humans. (a) Histogram of human answer accuracy (b) Center biases of the correct and incorrect attention}
    \label{fig:et_stats}
\end{figure}

\textbf{Human attention maps and performances.} Eye fixations are extracted from the raw data using the Cluster Fix algorithm~\cite{cluster_fix}, and a fixation map is computed for each question by aggregating the fixations from all participants. The fixation maps are scaled into $256\times 256$ pixels, smoothed using a Gaussian kernel ($\sigma=9$ pixels, $\approx 1$ DVA) and normalized to the range of [0,1]. The overall accuracy of human answers is $77.64\pm 24.55\%$ (M$\pm$SD). A total of 479 questions have consistently correct answers, and 934 have both correct and incorrect answers. The histogram of human answer accuracy is shown in 
\fig~\ref{fig:et_stats}a. We further separate the fixations into two groups based on answer correctness and compute a fixation map for each group. Correct and incorrect answers have comparable numbers of fixations per trial (10.12 \vs 10.27), while the numbers of fixations for the correct answers have a lower standard deviation across trials (0.99 \vs 1.54). \fig~\ref{fig:et_stats}b shows the prior distributions of the two groups of fixations, and their high similarity (Pearson's $r=0.997$) suggests that the answer correctness is independent of center bias. The correct and incorrect fixation maps are considered as two human attention baselines to compare with machine attentions, and also play a role in validating the effectiveness of the proposed AiR-E metric. More illustration is provided in the supplementary video.
\section{Experiments and Analyses} \label{analysis}

In this section, we conduct experiments and analyze various attention mechanisms of humans and machines. %More experiment details are provided in our supplementary materials. 
Our experiments aim to shed light on the following questions that have yet to be answered:
\begin{enumerate}
    \item Do machines or humans look at places relevant to the reasoning process? How does the attention process influence task performances? (Section~\ref{att_outcome})  %of humans and machines?
    \item How does attention accuracy evolve over time, and what about its correlation with the reasoning process? (Section~\ref{att_process})
    \item Does guiding models to look at places progressively following the reasoning process help? (Section~\ref{att_supervision}) 
\end{enumerate}

\subsection{Do Machines or Humans Look at Places Important to Reasoning? How Does Attention Influence Task Performances?} \label{att_outcome}
First, we measure attention accuracy throughout the reasoning process with the proposed AiR-E metric. Answer correctness is also compared, and its correlation with the attention accuracy reveals the joint influence of attention and reasoning operations to task performance. With these experiments, we observe that humans attend more accurately than machines, and the correlation between attention accuracy and task performance is dependent on the reasoning operations.

\begin{figure}[t]
\centering
\includegraphics[width=1\linewidth]{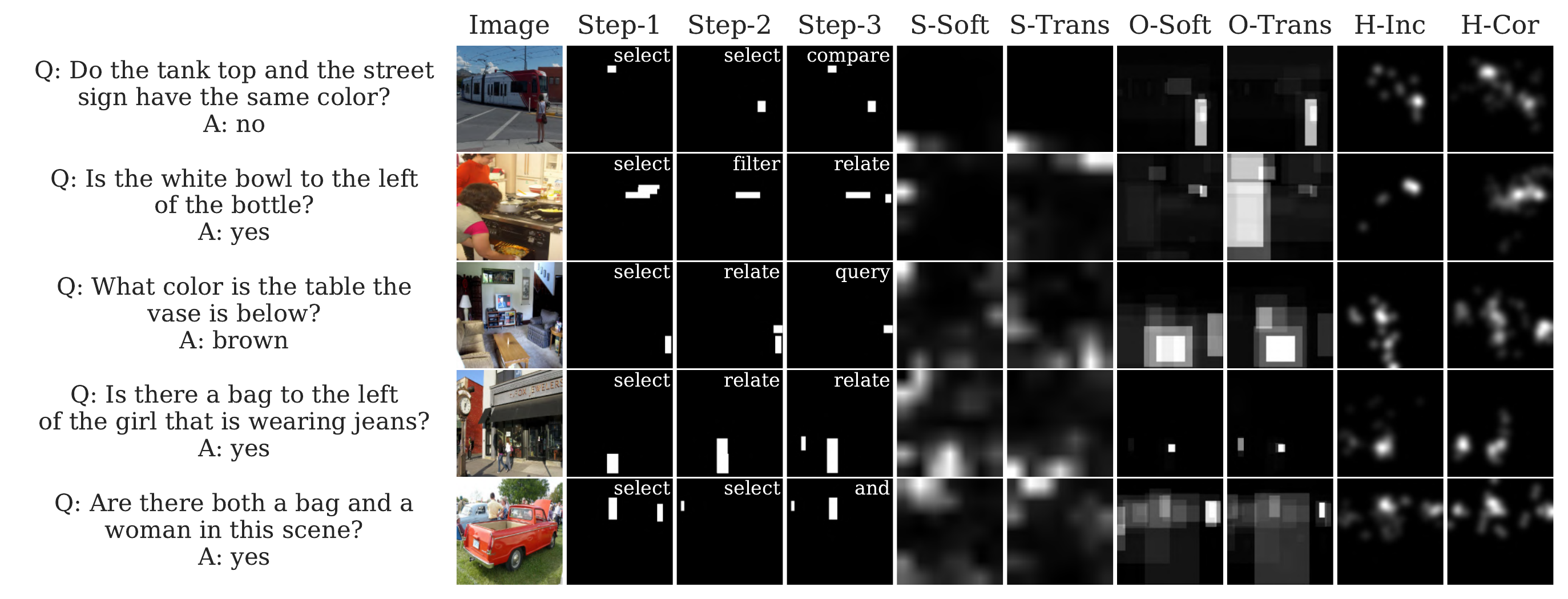}
\caption{Example question-answer pairs (column 1), images (column 2), ROIs at each reasoning step (columns 3-5), and attention maps (columns 6-11)}
\label{fig:qual}
\end{figure}

We evaluate four types of attentions that are commonly used in VQA models, including spatial soft attention (S-Soft), spatial Transformer attention (S-Trans), object-based soft attention (O-Soft), and object-based Transformer attention (O-Trans). Spatial and object-based attentions differ in terms of their inputs (\ie~image features or regional features), while soft and Transformer attention methods differ in terms of the computational methods of attention (\ie~with convolutional layers or matrix multiplication). We use spatial features extracted from ResNet-101 \cite{resnet} and object-based features from~\cite{updown} as the two types of inputs, and follow the implementations of \cite{updown} and \cite{dfaf} for the soft attention~\cite{soft_att} and Transformer attention~\cite{mul_att} computation, respectively. We integrate the aforementioned attentions with different state-of-the-art VQA models as backbones. Our observations are general and consistent across various backbones. In the following sections, we use the results on UpDown~\cite{updown} for illustration (results for the other backbones are provided in the supplementary materials). For human attentions, we denote the fixation maps associated with correct and incorrect answers as H-Cor and H-Inc, and the aggregated fixation map regardless of correctness is denoted as H-Tot. \fig~\ref{fig:qual} presents examples of ROIs for different reasoning operations and the compared attention maps. 

\textbf{Attention accuracy and task performance of humans and models.} \tab~\ref{AiR-E} quantitatively compares the AiR-E scores and VQA task performance across humans and models with different types of attentions. The task performance for models is the classification score of the correct answer, while the task performance for humans is the proportion of correct answers. Three clear gaps can be observed from the table:
\begin{enumerate*}
\item[(1)] Humans who answer correctly have significantly higher AiR-E scores than those who answer incorrectly. 
\item[(2)] Humans consistently outperform models in both attention and task performance. 
\item[(3)] Object-based attentions attend much more accurately than spatial attentions.
\end{enumerate*}
The low AiR-E of spatial attentions confirms the previous conclusion drawn from the VQA-HAT dataset~\cite{vqahat}. By constraining the visual inputs to a set of semantically meaningful objects, object-based attention typically increases the probabilities of attending to the correct ROIs. Between the two object-based attentions, the soft attention slightly outperforms its Transformer counterpart. Since the Transformer attentions explicitly learn the inter-object relationships, they perform better for logical operations (\ie~\textit{and}, \textit{or}). However, due to the complexity of the scenes and fewer parameters used~\cite{mul_att}, they do not perform as well as soft attention.
The ranks of different attentions are consistent with the intuition and literature, suggesting the effectiveness of the proposed AiR-E metric.

\textbf{Attention accuracy and task performance among different reasoning operations.} Comparing the different operations, \tab~\ref{AiR-E} shows that \textit{query} is the most challenging operation for models. Even with the highest attention accuracy among all operations, the task performance is the lowest. This is probably due to the inferior recognition capability of models compared with humans. To humans, `compare' is the most challenging in terms of task performance, largely because it often appears in complex questions that require close attention to multiple objects and thus take longer processing time. Since models can process multiple input objects in parallel, their performance is not highly influenced by the number of objects to look at.

\setlength{\tabcolsep}{0.45em}
\begin{table}[t]
\begin{center}
\caption{Quantitative evaluation of AiR-E scores and task performance}
\label{AiR-E}
\resizebox{0.75\linewidth}{!}{
\begin{tabular}{clllllllll}
\toprule
& Attention &             and &         compare &           filter &              or &           query &           relate &           select &          verify \\
\midrule
\parbox[t]{2mm}{\multirow{7}{*}{\rotatebox[origin=c]{90}{AiR-E}}} &H-Tot     &  2.197 &    2.669 &   2.810 & 2.429 &  3.951 &   3.516 &   2.913 &   3.629 \\
&H-Cor     &  2.258 &    2.717 &   2.925 & 2.529 &  4.169 &   3.581 &   2.954 &   3.580 \\
&H-Inc     &  1.542 &    1.856 &   1.763 & 1.363 &  2.032 &   2.380 &   1.980 &   2.512 \\
\cmidrule{2-10}
&O-Soft     &  1.334 &    \textbf{1.204} &   \textbf{1.518} & 1.857 & \textbf{3.241} &   \textbf{2.243} &   \textbf{1.586} &   2.091 \\
&O-Trans     &  \textbf{1.579} &    1.046 &   1.202 & \textbf{1.910} &  3.041 &   1.839 &   1.324 &   \textbf{2.228} \\
&S-Soft     & -0.001 &   -0.110 &   0.251 & 0.413 &  0.725 &   0.305 &   0.145 &   0.136 \\
&S-Trans     &  0.060 &   -0.172 &   0.243 & 0.343 &  0.718 &   0.370 &   0.173 &   0.101 \\
\midrule
\parbox[t]{2mm}{\multirow{5}{*}{\rotatebox[origin=c]{90}{Accuracy}}}&H-Tot     & 0.700 &    0.625 &   0.668 & 0.732 &  0.633 &   0.672 &   0.670 &   0.707 \\
\cmidrule{2-10}
&O-Soft     & 0.604 &    \textbf{0.547} &   0.603 & 0.809 &  \textbf{0.287} &   0.483 &   0.548 &   0.605 \\
&O-Trans     & \textbf{0.606} &    0.536 &   \textbf{0.608} & \textbf{0.832} &  0.282 &   \textbf{0.487} &   \textbf{0.550} &   0.592 \\
&S-Soft     & 0.592 &    0.520 &   0.558 & 0.814 &  0.203 &   0.427 &   0.511 &   0.544 \\
&S-Trans     & 0.597 &    0.525 &   0.557 & 0.811 &  0.211 &   0.435 &   0.517 &   \textbf{0.607} \\

\bottomrule
\end{tabular}
}
\end{center}
\end{table} 

\begin{table}[t]
\begin{center}
\caption{Pearson's $r$ between attention accuracy (AiR-E) and task performance. Bold numbers indicate significant positive correlations (p$<$0.05)}
\label{corr}
\resizebox{0.75\linewidth}{!}{
\begin{tabular}{lllllllll}
\toprule
Attention &             and &         compare &           filter &              or &           query &           relate &           select &          verify \\
\midrule
H-Tot     &       0.205 &  \textbf{0.329} &         0.051 &  0.176 &  \textbf{0.282} &   \textbf{0.210} &   \textbf{0.134} &  \textbf{0.270} \\
\midrule
O-Soft     &       0.167 &  \textbf{0.217} &        -0.022 &  0.059 &  \textbf{0.331} &         0.058 &         0.003 &        0.121 \\
O-Trans     &       0.168 &  \textbf{0.205} &         0.090 &  0.174 &  \textbf{0.298} &         0.041 &   \textbf{0.063} &       -0.027 \\
S-Soft     &       0.177 &  \textbf{0.237} &        -0.084 &  0.082 &       -0.017 &  -0.170 &  -0.084 &        0.066 \\
S-Trans     &       0.171 &  \textbf{0.210} &  -0.152 &  0.086 &       -0.024 &  -0.139 &  -0.100 &  \textbf{0.270} \\
\bottomrule
\end{tabular}
}
\end{center}
\end{table} 

\textbf{Correlation between attention accuracy and task performance.} The similar rankings of AiR-E and task performance suggest a correlation between attention accuracy and task performance. To further investigate this correlation on a sample basis, for each attention and operation, we compute the Pearson's $r$ between the attention accuracy and task performance across different questions. 

As shown in \tab~\ref{corr}, human attention accuracy and task performance are correlated for most of the operations (up to $r=0.329$). The correlation is higher than most of the compared machine attentions, suggesting that humans' task performance is more consistent with their attention quality. In contrast, though commonly referred as an interface for interpreting models' decisions~\cite{vqahat,explanation,qual_eval}, spatial attention maps do not reflect the decision-making process of models. They typically have very low and even negative correlations (\eg~\textit{relate}, \textit{select}). By limiting the visual inputs to foreground objects, object-based attentions achieve higher attention-answer correlations.

The differences of correlations between operations are also significant. For the questions requiring focused attention to answer (\ie~with \textit{query} and \textit{compare} operations), the correlations are relatively higher than the others. 

\subsection{How Does Attention Accuracy Evolve Throughout the Reasoning Process?} \label{att_process}
\begin{figure}[t]
\centering
    \subfloat[]{\includegraphics[width=0.32\linewidth]{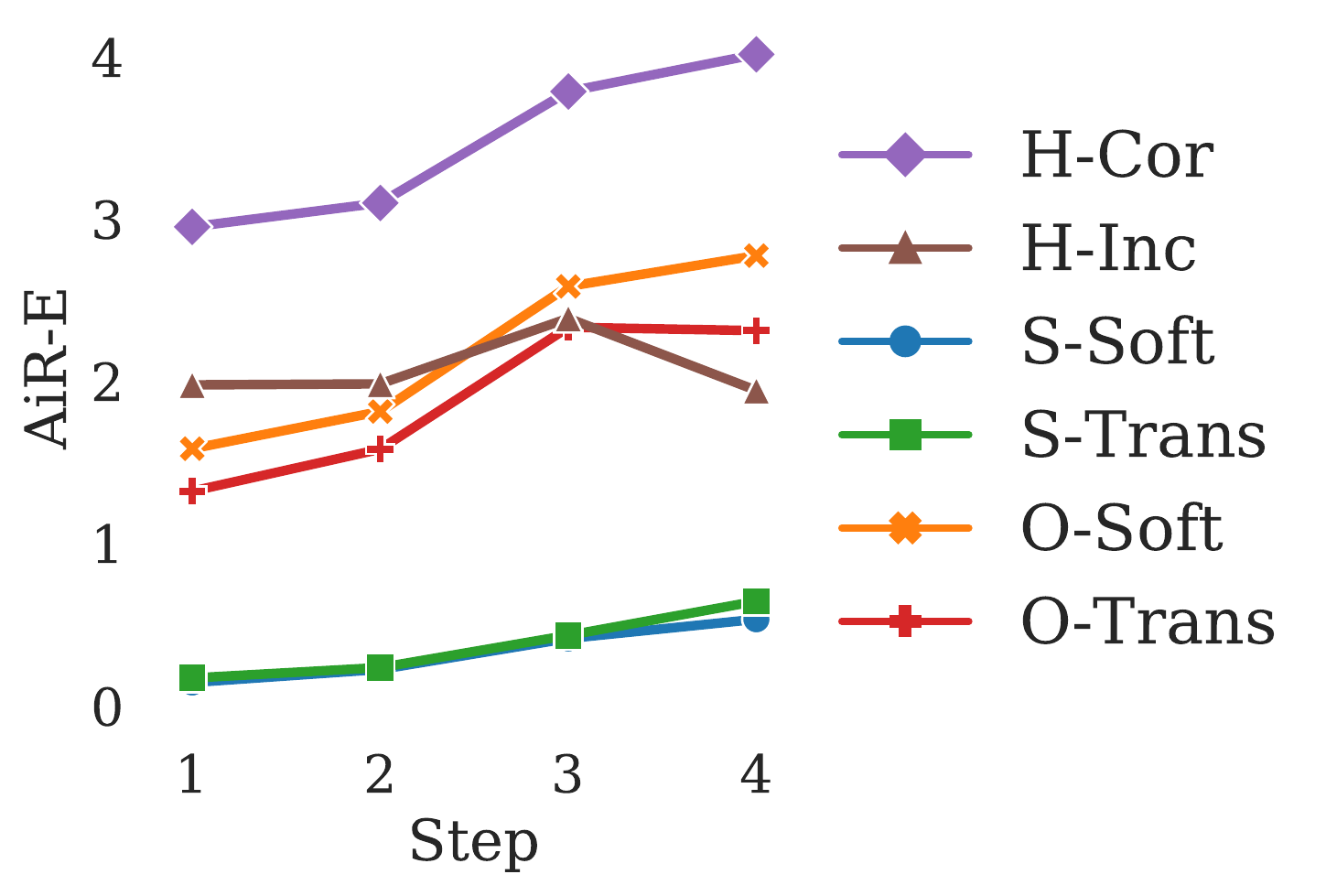}}%'height=8cm' is needed for this example only and can be dropped when using it with actual image
     \vrule
     \hfill
    \subfloat[]{\includegraphics[width=0.275\linewidth]{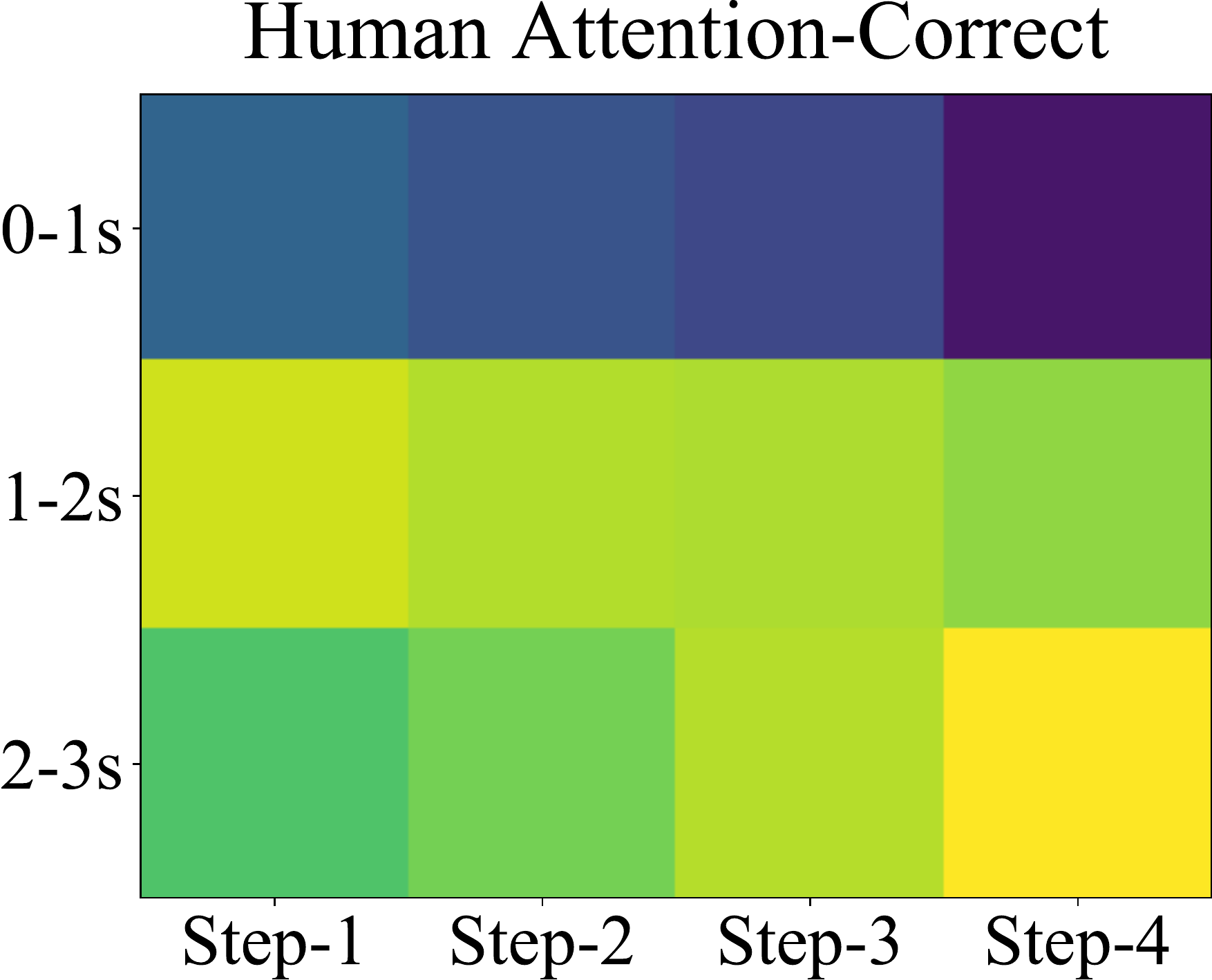}}
    \hfill
    \subfloat[]{\includegraphics[width=0.32\linewidth]{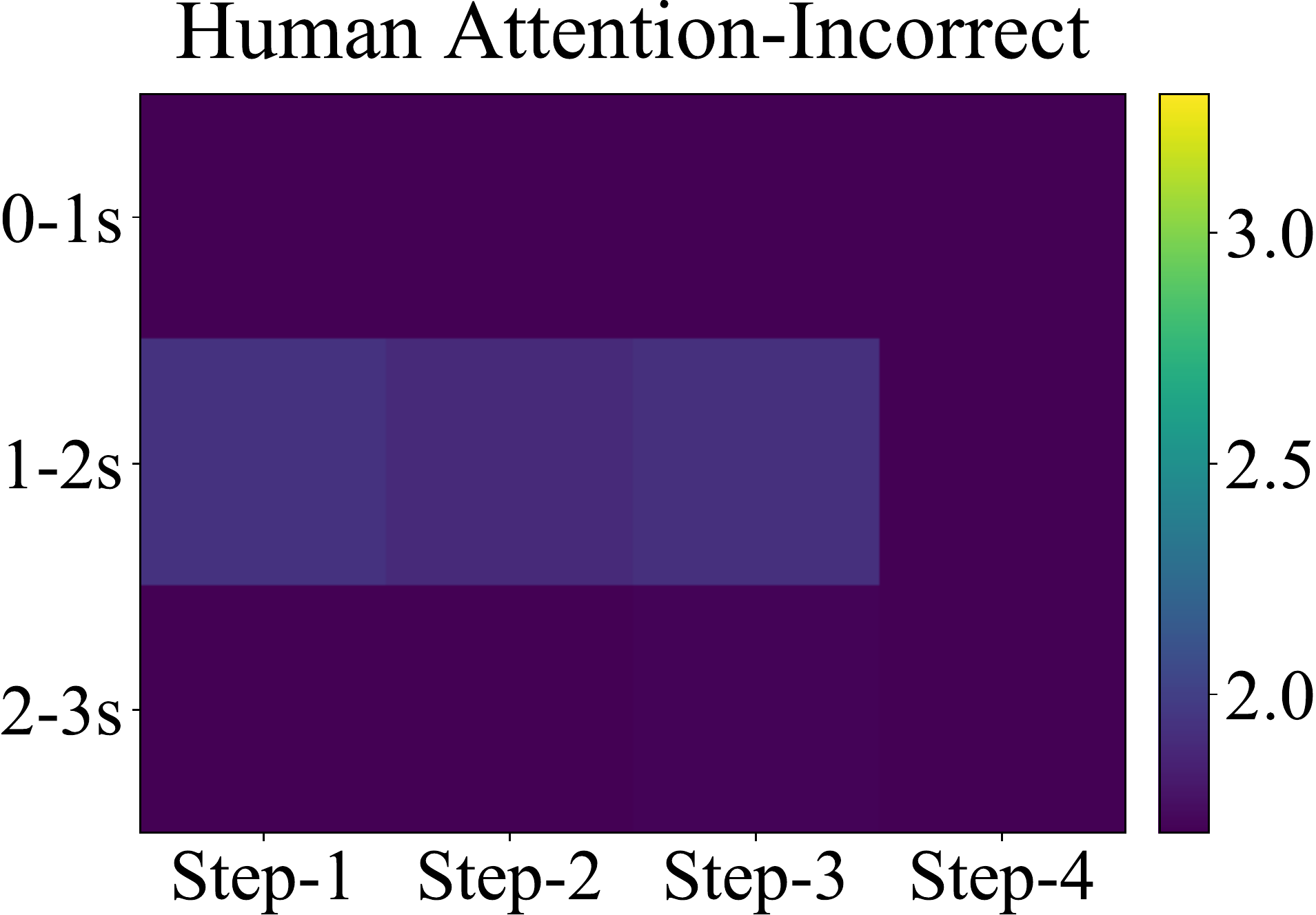}}
    \hfill
    \\
    \hrulefill\\
    \subfloat[]{\includegraphics[width=0.275\linewidth]{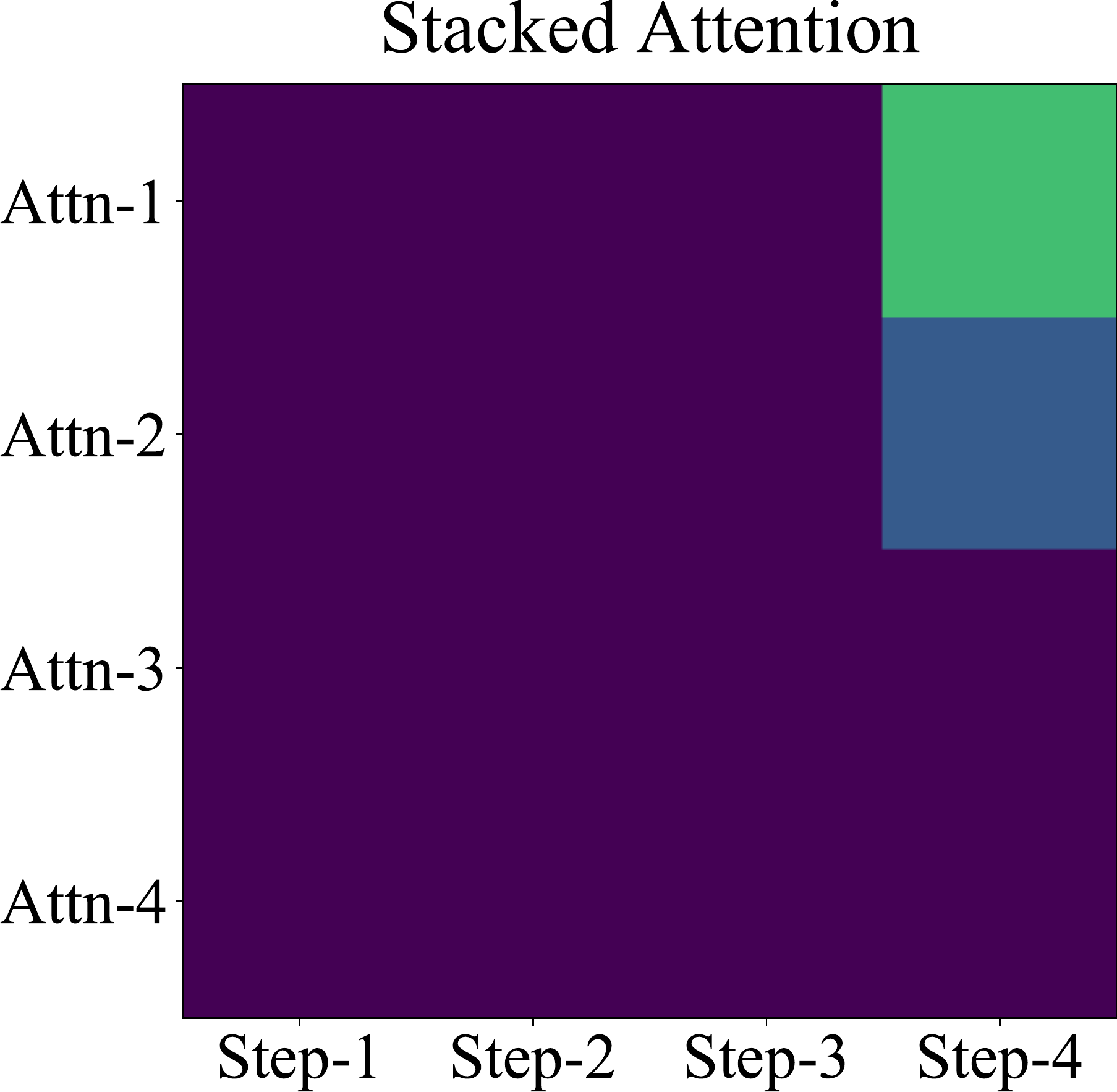}}
    \hfill
    \subfloat[]{\includegraphics[width=0.275\linewidth]{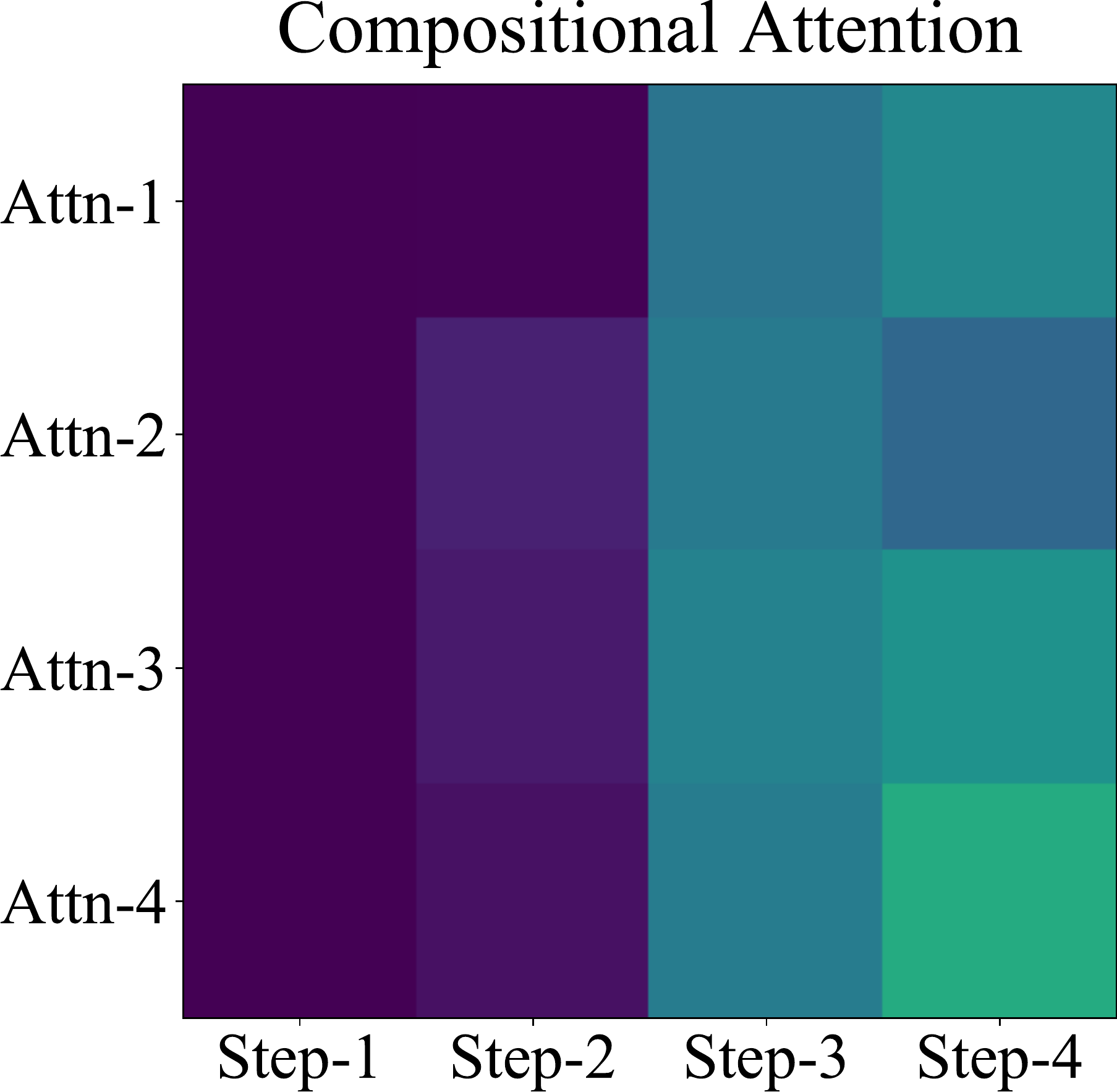}}
    \hfill
\subfloat[]{\includegraphics[width=0.32\linewidth]{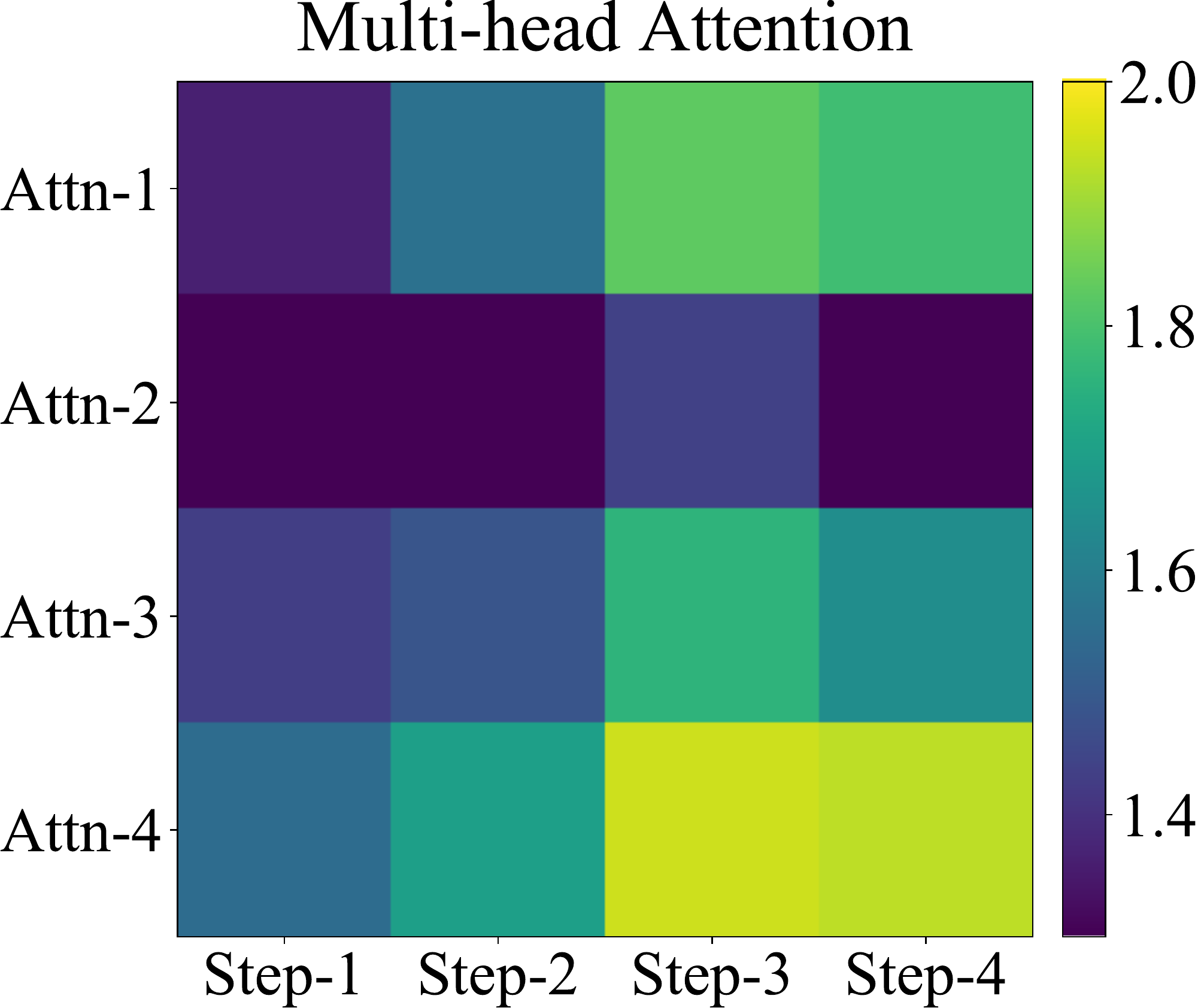}}       
  
  \caption{Spatiotemporal accuracy of attention throughout the reasoning process. (a) shows the AiR-E of different reasoning steps for human aggregated attentions and single-glimpse machine attentions, (b)-(c) AiR-E scores for decomposed human attentions with correct and incorrect answers, (d)-(f) AiR-E for multi-glimpse machine attentions. For heat maps shown in (b)-(f), the x-axis denotes different reasoning steps while the y-axis corresponds to the indices of attention maps} 
  \label{fig:step_comparison}
\end{figure}

To complement our previous analysis on the spatial allocation of attentions,we move forward to analyze the spatiotemporal alignment of attentions. Specifically, we analyze the AiR-E scores according to the chronological order of reasoning operations. We show in \fig~\ref{fig:step_comparison}a that the AiR-E scores peak at the $3^{rd}$ or $4^{th}$ steps, suggesting that human and machine attentions focus more on the ROIs closely related to the final task outcome, instead of the earlier steps. In the rest of this section, we focus our analysis on the spatiotemporal alignment between multiple attention maps and the ROIs at different reasoning steps. In particular, we study the change of human attention over time, and compare it with multi-glimpse machine attentions. Our analysis reveals the significant spatiotemporal discrepancy between human and machine attentions.

\textbf{Do human attentions follow the reasoning process?}
First, to analyze the spatiotemporal deployment of human attention in visual reasoning, we group the fixations into three temporal bins (0-1s, 1-2s and 2-3s), and compute AiR-E scores for each fixation map and reasoning step (see \fig~\ref{fig:step_comparison}b-c). Humans start exploration (0-1s) with relatively low attention accuracy. After the initial exploration, human attention shows improved accuracy across all reasoning steps (1-2s), and particularly focuses on the early-step ROIs. In the final steps (2-3s), depending on the correctness of answers, human attention either shifts to the ROIs at later stages (correct), or becomes less accurate with lowered AiR-E scores (incorrect). Such observations suggest high spatiotemporal alignments between human attention and the sequence of reasoning operations. 

\textbf{Do machine attentions follow the reasoning process?}
Similarly, we evaluate multi-glimpse machine attentions. We compare the stacked attention from SAN~\cite{san}, compositional attention from MAC~\cite{mac} and the multi-head attention~\cite{mcb,mfb}, which all adopt the object-based attention. \fig~\ref{fig:step_comparison}d-f shows that multi-glimpse attentions do not evolve with the reasoning process. Stacked attention's first glimpse already attends to the ROIs at the $4^{th}$ step, and the other glimpses contribute little to the attention accuracy. Compositional attention and multi-head attention consistently align the best with the ROIs at the $3^{rd}$ or $4^{th}$ step, and ignore those at the early steps.

The spatiotemporal correlations indicate that following the correct order of reasoning operations is important for humans to attend and answer correctly. In contrast, models tend to directly attend to the final ROIs, instead of shifting their attentions progressively.
\begin{table}[t]
\begin{center}
\caption{Comparative results on GQA test sets (test-dev and test-standard). We report the single-model performance trained on the balanced training set of GQA}
\label{model}
\resizebox{0.75\linewidth}{!}{
\begin{tabular}{c|c|c|c|c|c|c}
\hline
&  \multicolumn{2}{c|}{UpDown \cite{updown}} & \multicolumn{2}{c|}{MUTAN \cite{mutan}} & \multicolumn{2}{c}{BAN \cite{ban}}\\
\cline{2-7}
  & dev  & standard & dev  & standard & dev  & standard   \\
\hline
w/o Supervision & 51.31 & 52.31 & 50.78 & 51.16 & 50.14 & 50.38 \\
PAAN \cite{paan} & 48.03 & 48.92 & 46.40 & 47.22 & n/a & n/a \\ HAN \cite{han} & 49.96 & 50.58 & 48.76 & 48.99 & n/a & n/a\\
ASM \cite{mining} & 52.96 & 53.57 & 51.46 & 52.36 & n/a & n/a\\ 
AiR-M & \textbf{53.46} & \textbf{54.10} & \textbf{51.81} & \textbf{52.42} & \textbf{53.36} & \textbf{54.15} \\
\hline
\end{tabular}
}
\end{center}
\end{table}
\begin{figure}
\centering
\includegraphics[width=0.95\linewidth]{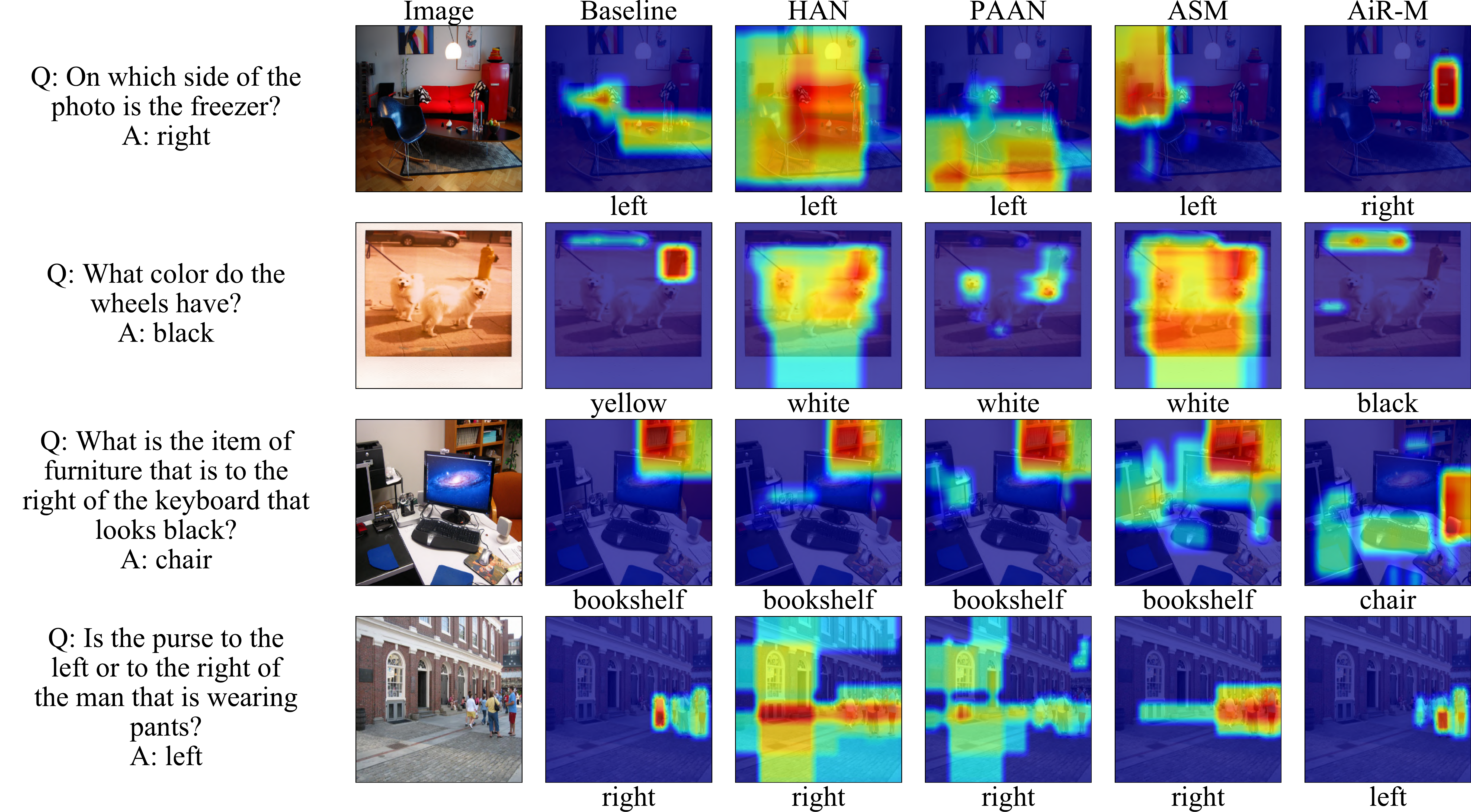}
\caption{Qualitative comparison between attention supervision methods, where Baseline refers to UpDown~\cite{updown}. For each row, from left to right are the questions and the correct answers, input images, and attention maps learned by different methods. The predicted answers associated with the attentions are shown below its respective attention map}
\label{fig:model_qualitative}
\end{figure}

\subsection{Does Progressive Attention Supervision Improve Attention and Task Performance?} \label{att_supervision}
Experiments in Section \ref{att_outcome} and Section \ref{att_process} suggest that attention towards ROIs relevant to the reasoning process contributes to task performance, and furthermore, the order of attention matters. Therefore, we propose to guide models to look at places important to reasoning in a progressive manner. Specifically, we propose to supervise machine attention along the reasoning process by jointly optimizing attention, reasoning operations, and task performance (AiR-M, see Section \ref{sup_method}). Here we investigate the effectiveness of the AiR-M supervision method on three VQA models, \ie~UpDown~\cite{updown}, MUTAN~\cite{mutan}, and BAN~\cite{ban}. We compare AiR-M with a number of state-of-the-art attention supervision methods, including supervision from human-like attention (HAN)~\cite{han}, attention supervision mining (ASM)~\cite{mining} and adversarial learning (PAAN)~\cite{paan}. Note that while the other compared methods are typically limited to supervision on a single attention map, our AiR-M method is generally applicable to various VQA models with single or multiple attention maps (\eg~BAN \cite{ban}). %The experimental results demonstrate the advantages of AiR-M, which is able to simultaneously improve the attention accuracy and task performance with broad generalizability.

According to \tab~\ref{model}, the proposed AiR-M supervision significantly improves the performance of all baselines and consistently outperforms the other attention supervision methods. Two of the compared methods, HAN and PAAN, fail to improve the performance of object-based attention. Supervising attention with knowledge from objects mined from language, ASM~\cite{mining} is able to consistently improve the performance of models. However, without considering the intermediate steps of reasoning, it is not as effective as the proposed method. 

\fig~\ref{fig:model_qualitative} shows the qualitative comparison between supervision methods. The proposed AiR-M not only directs attention to the ROIs most related to the answers (\ie~freezer, wheel, chair, purse), but also highlights other important ROIs mentioned in the questions (\ie~ keyboard, man), thus reflecting the entire reasoning process, while attentions in other methods fail to localize these ROIs.

\begin{figure}[t]
\centering
\includegraphics[width=0.3\linewidth]{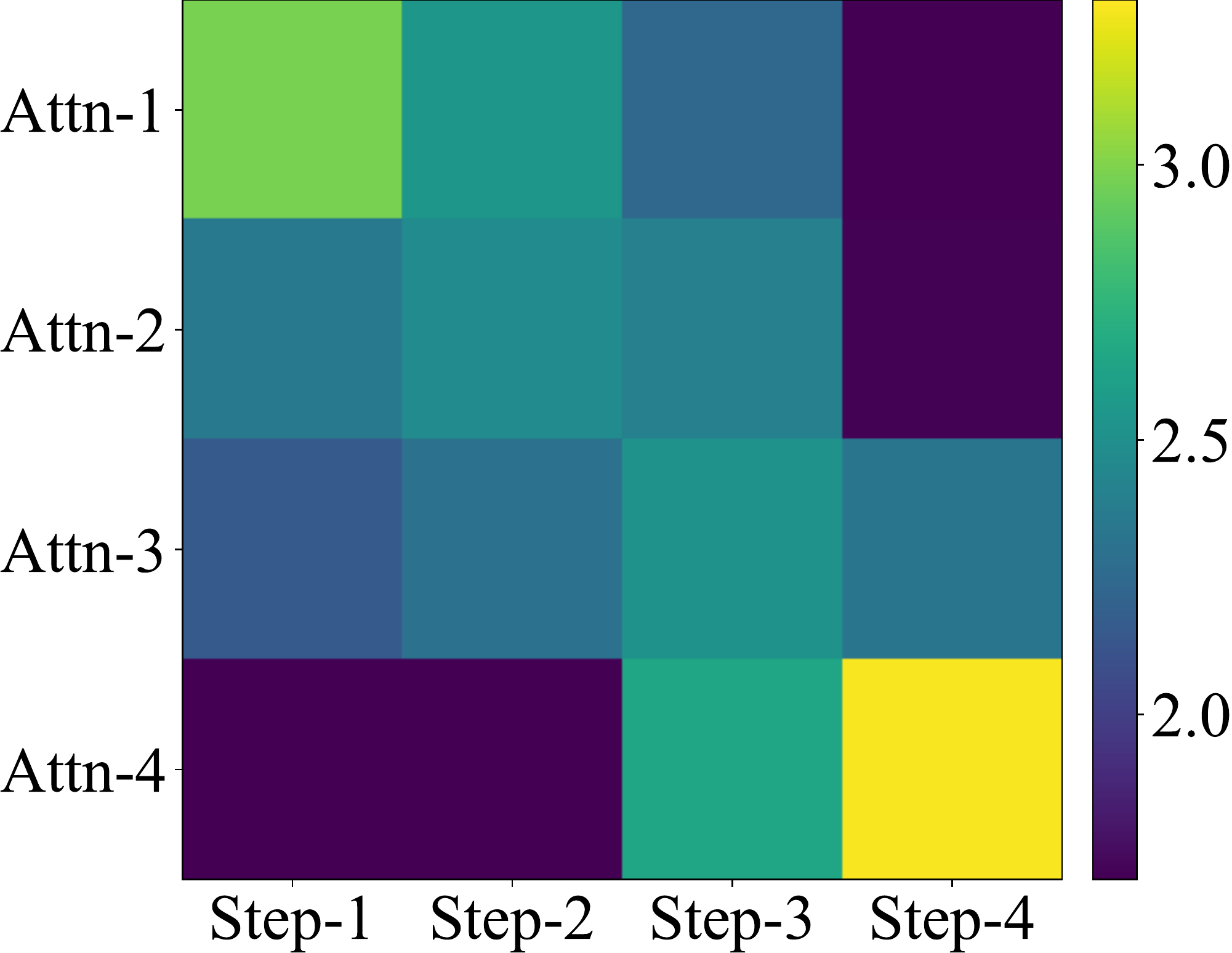}
\caption{Alignment between the proposed attention and reasoning process}
\label{fig:ours}
\vspace{-1em}
\end{figure}

\tab~\ref{corr_supervision} reports the AiR-E scores across operations. It shows that the AiR-M supervision method significantly improves attention accuracy (attention aggregated across different steps), especially on those typically positioned in early steps (\eg~\textit{select}, \textit{compare}). In addition, the AiR-M supervision method also aligns the multi-glimpse attentions better according to their chronological order in the reasoning process (see \fig~\ref{fig:ours} and the supplementary video), showing progressive improvement of attention throughout the entire process.  
\begin{table}[t]
\begin{center}
\caption{AiR-E scores of the supervised attentions}
\label{corr_supervision}
\resizebox{0.75\linewidth}{!}{
\begin{tabular}{lllllllll}
\toprule
Attention &             and &         compare &           filter &              or &           query &           relate &           select &          verify \\
\midrule
Human   &  2.197 &    2.669 &   2.810 & 2.429 &  3.951 &   3.516 &   2.913 &   3.629\\
\midrule
AiR-M       & \textbf{2.396} &    \textbf{2.553} &   \textbf{2.383} & \textbf{2.380} &  3.340 &   \textbf{2.862} &   \textbf{2.611} &   \textbf{4.052} \\
Baseline \cite{updown} & 1.859 &    1.375 &   1.717 & 2.271 &  \textbf{3.651} &   2.448 &   1.796 &   2.719 \\
ASM    & 1.415 &    1.334 &   1.443 & 1.752 &  2.447 &   1.884 &   1.584 &   2.265 \\
HAN       & 0.581 &    0.428 &   0.468 & 0.607 &  1.576 &   0.923 &   0.638 &   0.680 \\
PAAN      & 1.017 &    0.872 &   1.039 & 1.181 &  2.656 &   1.592 &   1.138 &   1.221 \\
\bottomrule
\end{tabular}
}
\end{center}
\end{table} 

\section{Conclusion}
We introduce AiR, a novel framework with a quantitative evaluation metric (AiR-E), a supervision method (AiR-M), and an eye-tracking dataset (AiR-D) for understanding and improving attention in the reasoning context. % AiR enables fine-grained analyses of human and machine attention from different aspects. 
Our analyses show that accurate attention deployment can lead to improved task performance, which is related to both the task outcome and the intermediate reasoning steps. Our experiments also highlight the significant gap between models and humans on the alignment of attention and reasoning process. With the proposed attention supervision method, we further demonstrate that incorporating the progressive reasoning process in attention can improve the task performance by a considerable margin. We hope that this work will be helpful for future development of visual attention and reasoning method, and inspire the analysis of model interpretability throughout the decision-making process.

\section*{Acknowledgements}
This work is supported by NSF Grants 1908711 and 1849107.

\clearpage

% \bibliography{egbib}

\clearpage

\title{AiR: Attention with Reasoning Capability (Supplementary Materials)}
\author{Shi Chen\thanks{Equal contributions.}\orcidID{0000-0002-3749-4767} \and
Ming Jiang\inst{\star}\orcidID{0000-0001-6439-5476} \and
Jinhui Yang\orcidID{0000-0001-8322-1121} \and
Qi Zhao\orcidID{0000-0003-3054-8934}}
\authorrunning{S. Chen et al.}
\institute{
University of Minnesota, Minneapolis MN 55455, USA\\
\email{\{chen4595,mjiang,yang7004,qzhao\}@umn.edu}}

\maketitle
The supplementary materials consist of results and details of the proposed Attention with Reasoning capability (AiR) framework:
\begin{enumerate}
    \item We complement the results presented in the main paper with analyses on different baselines, including Multi-modal Factorized Bilinear (MFB) \cite{mfb} and Multi-Modal Tucker Fusion (MUTAN) \cite{mutan} (Section~\ref{sec:supp_analyses}).
    \item We present ablation studies on different attention supervision strategies and hyperparameters of the proposed AiR-M method  (Section~\ref{sec:supp_ablation} and Section~\ref{sec:supp_ablation_para}).
    \item We present additional qualitative comparisons between the AiR-M method and various attention supervision methods (Section~\ref{sec:supp_qual}).
    \item We present details about the decomposition of the reasoning process (Section~\ref{sec:supp_process}).
    \item We report details of the proposed AiR-M method with applications to different types of attention mechanisms (Section~\ref{sec:supp_supervision}).
    % \item We present additional results on the proposed attention supervision method (AiR-M).
\end{enumerate}

We also provide a \textbf{supplementary video} to illustrate the spatiotemporal dynamics of both model and human attentions throughout the reasoning process. It demonstrates the effectiveness of the proposed attention supervision method (AiR-M) in improving both the attention accuracy and task performance. % It demonstrates the superiority of the proposed attention supervision method (AiR-M) on learning attentions desirable for different reasoning steps. Compared to other methods \cite{paan,han,mining} that either answer correctly without attending to the correct ROIs or answer incorrectly due to the failure of capturing the essential ROIs, our method is able to progressively attend to ROIs throughout the reasoning process and answer with high accuracy. 
In addition, the video also highlights the significant spatiotemporal discrepancy between human attentions with correct and incorrect answers. 

\section{Supplementary Results}

\subsection{Analyses on MFB and MUTAN Models}
\label{sec:supp_analyses}
Due to the page limit of the main paper, here we provide supplementary analyses on two additional baselines, \ie~MFB \cite{mfb} and MUTAN \cite{mutan}, to demonstrate the generality of the analyses for different baselines. %, we present supplementary results on two additional VQA models, MFB \cite{mfb} and MUTAN \cite{mutan}. 
The experimental procedures are consistent with those in Section 4.1 of main paper. 

\setlength{\tabcolsep}{0.45em}
\begin{table}
\begin{center}
\caption{Quantitative evaluation of AiR-E scores and task performances of the MFB \cite{mfb} baseline. Bold numbers indicate the best attention performance.}
\label{nos_mfb}
\resizebox{0.8\linewidth}{!}{
\begin{tabular}{clllllllll}
\toprule
& Attention &             and &         compare &           filter &              or &           query &           relate &           select &          verify \\
\midrule
\parbox[t]{2mm}{\multirow{7}{*}{\rotatebox[origin=c]{90}{AiR-E}}} &H-Tot &  2.197 &  2.669 &   2.810 & 2.429 &  3.951 &   3.516 &   2.913 &   3.629 \\
&H-Cor     &  2.258 &    2.717 &   2.925 & 2.529 &  4.169 &   3.581 &   2.954 &   3.580 \\
&H-Inc     &  1.542 &    1.856 &   1.763 & 1.363 &  2.032 &   2.380 &   1.980 &   2.512 \\
\cmidrule{2-10}
&MFB-O-Soft     &  \textbf{1.841}  & 1.055  &  \textbf{1.294}  & \textbf{2.295} & \textbf{3.799}  &  \textbf{1.779} &  1.372  & \textbf{2.563}  \\
&MFB-O-Trans     &  1.787  &  \textbf{1.446} &  1.054  &  1.957 &  3.740 & 1.730  & \textbf{1.405}   &  2.386  \\
&MFB-S-Soft     & 0.217 & -0.044 & 0.176  & 0.477 & 0.746  & 0.341  & 0.195   & 0.005 \\
&MFB-S-Trans     & 0.438 &  0.367  & 0.524  & 0.702 & 0.765  & 0.640  & 0.468   &  0.652  \\
\midrule
\parbox[t]{2mm}{\multirow{5}{*}{\rotatebox[origin=c]{90}{Accuracy}}}&H-Tot     & 0.700 &    0.625 &   0.668 & 0.732 &  0.633 &   0.672 &   0.670 &   0.707 \\
\cmidrule{2-10}
&MFB-O-Soft     &  0.626 &  0.593  & 0.549  & \textbf{0.834} & \textbf{0.389}  & \textbf{0.467}  &  \textbf{0.533}   & 0.645 \\
&MFB-O-Trans     & \textbf{0.631} &  \textbf{0.598}  & \textbf{0.550}  &  0.833 &  0.388 & 0.466  & \textbf{0.533}  & \textbf{0.649} \\
&MFB-S-Soft     & 0.595 &  0.581  & 0.508  & 0.805 & 0.316  & 0.408  & 0.481  & 0.614 \\
&MFB-S-Trans     & 0.598 &  0.581  &  0.506  & 0.802  & 0.313  & 0.406  & 0.479  & 0.615 \\
\bottomrule
\end{tabular}
}
\end{center}
% \vspace{-4em}
\end{table} 

\begin{table}
\begin{center}
\caption{Pearson's correlation coefficients between attention accuracy (AiR-E) and task performances of the MFB \cite{mfb} baseline. Bold numbers indicate significant positive correlations (p$<$0.05).}
\label{corr_mfb}
\resizebox{0.8\linewidth}{!}{
\begin{tabular}{lllllllll}
\toprule
Attention &             and &         compare &           filter &              or &           query &           relate &           select &          verify \\
\midrule
H-Tot     &       0.205 &  \textbf{0.329} &         0.051 &  0.176 &  \textbf{0.282} &   \textbf{0.210} &   \textbf{0.134} &  \textbf{0.270} \\
MFB-O-Soft    &        0.103 &  -0.096 &   \textbf{0.131} &  \textbf{0.244} &  \textbf{0.370} &         0.045 &         0.041 &  \textbf{0.182} \\
MFB-O-Trans   &  \textbf{0.225} &   0.050 &   \textbf{0.121} &  \textbf{0.197} &  \textbf{0.370} &         0.038 &         0.043 &  \textbf{0.256} \\
MFB-S-Soft    &        0.027 &   0.064 &        -0.084 &        0.015 &  \textbf{0.219} &        -0.013 &        -0.028 &        0.084 \\
MFB-S-Trans   &       -0.013 &   0.077 &        -0.033 &  \textbf{0.238} &  \textbf{0.165} &         0.037 &         0.001 &       -0.019 \\
\bottomrule
\end{tabular}
}
\end{center}
% \vspace{-4em}
\end{table} 

\setlength{\tabcolsep}{0.45em}
\begin{table}
\begin{center}
\caption{Quantitative evaluation of AiR-E scores and task performances of the MUTAN \cite{mutan} baseline. Bold numbers indicate the best attention performance.}
\label{nos_mutan}
\resizebox{0.8\linewidth}{!}{
\begin{tabular}{clllllllll}
\toprule
& Attention &             and &         compare &           filter &              or &           query &           relate &           select &          verify \\
\midrule
\parbox[t]{2mm}{\multirow{7}{*}{\rotatebox[origin=c]{90}{AiR-E}}} &H-Tot &  2.197 &  2.669 &   2.810 & 2.429 &  3.951 &   3.516 &   2.913 &   3.629 \\
&H-Cor     &  2.258 &    2.717 &   2.925 & 2.529 &  4.169 &   3.581 &   2.954 &   3.580 \\
&H-Inc     &  1.542 &    1.856 &   1.763 & 1.363 &  2.032 &   2.380 &   1.980 &   2.512 \\
\cmidrule{2-10}
&MUTAN-O-Soft     &  \textbf{2.051}  & \textbf{1.490}  &  \textbf{1.676}  &  \textbf{2.644} & \textbf{3.683}  &  \textbf{2.096} &  \textbf{1.695}  & \textbf{2.762}  \\
&MUTAN-O-Trans     &  0.973  &  0.851 &  1.137  &  1.655 &  2.559  & 1.609  & 1.130   &  1.974  \\
&MUTAN-S-Soft     & 0.124 & 0.098 & 0.253  &  0.347 & 0.761  & 0.359  & 0.243   &  0.172 \\
&MUTAN-S-Trans     & 0.290 &  0.074  & 0.208  & 0.182 & 0.778  & 0.399  & 0.253   &  0.155  \\
\midrule
\parbox[t]{2mm}{\multirow{5}{*}{\rotatebox[origin=c]{90}{Accuracy}}}&H-Tot     & 0.700 &    0.625 &   0.668 & 0.732 &  0.633 &   0.672 &   0.670 &   0.707 \\
\cmidrule{2-10}
&MUTAN-O-Soft     &  \textbf{0.602} &  \textbf{0.593}  & \textbf{0.541}  & \textbf{0.787} & \textbf{0.385}  & \textbf{0.457}  &  \textbf{0.521}   & \textbf{0.645} \\
&MUTAN-O-Trans     & 0.582 &  0.588  & 0.529  &  0.771 &  0.374  & 0.443  & 0.507  & 0.635 \\
&MUTAN-S-Soft     & 0.568 &  0.583  & 0.502  & 0.765 & 0.320  & 0.404  & 0.473  & 0.616 \\
&MUTAN-S-Trans     & 0.576 &  0.583  &  0.501  & 0.763  & 0.319  & 0.402  & 0.473  & 0.615 \\
\bottomrule
\end{tabular}
}
\end{center}
% \vspace{-4em}
\end{table} 

\begin{table}
\begin{center}
\caption{Pearson's correlation coefficients between attention accuracy (AiR-E) and task performance of the MUTAN \cite{mutan} baseline. Bold numbers indicate significant positive correlations (p$<$0.05).}
\label{corr_mutan}
\resizebox{0.8\linewidth}{!}{
\begin{tabular}{lllllllll}
\toprule
Attention &             and &         compare &           filter &              or &           query &           relate &           select &          verify \\
\midrule
H-Tot     &       0.205 &  \textbf{0.329} &         0.051 &  0.176 &  \textbf{0.282} &   \textbf{0.210} &   \textbf{0.134} &  \textbf{0.270} \\
MUTAN-O-Soft  &        0.142 &  -0.027 &   \textbf{0.130} &  \textbf{0.205} &  \textbf{0.369} &         0.034 &         0.018 &  \textbf{0.229} \\
MUTAN-O-Trans &  \textbf{0.174} &   0.027 &         0.015 &        0.128 &  \textbf{0.284} &   \textbf{0.076} &         0.002 &        0.122 \\
MUTAN-S-Soft  &  \textbf{0.228} &   0.062 &  {-0.160} &        0.059 &  \textbf{0.121} &  {-0.090} &  {-0.058} &        0.116 \\
MUTAN-S-Trans &       -0.021 &   0.065 &  {-0.208} &       -0.034 &  \textbf{0.105} &  {-0.064} &  {-0.118} &        0.074 \\
\bottomrule
\end{tabular}
}
\end{center}
% \vspace{-2em}
\end{table} 

\begin{figure}[t]
\centering
    \includegraphics[width=0.4\linewidth]{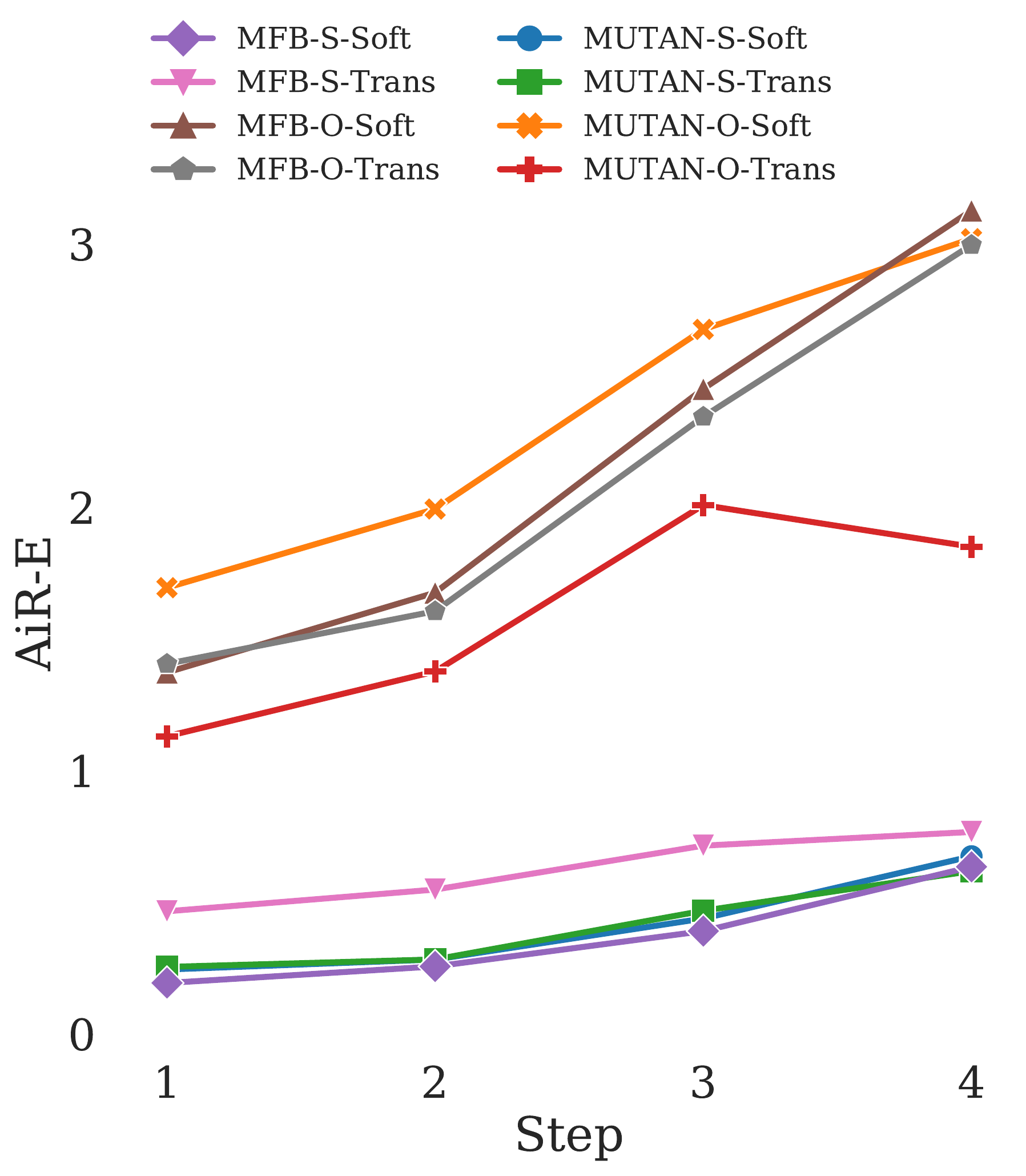}   
  
  \caption{Spatiotemporal accuracy of attention throughout the reasoning process.} 
  \label{fig:step_comparison_supp}
  \vspace{-1em}
\end{figure}

% \clearpage

Overall, the results on MFB \cite{mfb} and MUTAN \cite{mutan} agree with our observations reported in the main paper (on UpDown~\cite{updown}): 
\begin{enumerate*}
    \item[(1)] Compared with computational models, humans have stronger reliance on attention and attend more accurately during visual reasoning (see Table~\ref{nos_mfb} and Table~\ref{nos_mutan}), which agrees with Table 2 of the main paper;
    \item[(2)] The task performance is jointly influenced by both attention accuracy and reasoning operations (see Table~\ref{corr_mfb} and Table~\ref{corr_mutan}), which agrees with Table 3 of the main paper;
    \item[(3)] Existing machine attentions tend to focus more on the ROIs closely related to the task outcome but are less related to intermediate reasoning steps (see Fig.~\ref{fig:step_comparison_supp}), which agrees with Fig.~5a of the main paper.
\end{enumerate*}
These agreements confirm that our observations in the main paper are general and consistent across different baselines. 

\subsection{Ablation Study of Attention Supervision Strategies}
\label{sec:supp_ablation}

To evaluate the effectiveness of different components in the proposed AiR-M attention supervision method, we conduct an ablation study on the UpDown \cite{updown} baseline with different supervision strategies:
\begin{enumerate}
    \item Joint supervision of attention, reasoning operation, and task performance, but in a single-glimpse manner, denoted as $L_{\Vec{\alpha}}$ + $L_{\Vec{r}}$. Specifically, the model only predicts a single attention map by incorporating the last hidden state of the GRU for operation prediction, and attention supervision is accomplished with ground truth attention aggregated across all steps.
    \item Progressive supervision only on the reasoning operation and task performance, denoted as AiR-M (w/o $L_{\Vec{\alpha}}$).
    \item Progressive supervision only on attention and task performance, denoted as AiR-M (w/o $L_{\Vec{r}}$).
\end{enumerate}

\begin{table}
\begin{center}
% \scriptsize
\caption{Experimental results of AiR-M under different supervision strategies. All reported results are on the GQA \cite{gqa} test-dev set. Bold numbers indicate the best performance.}
\label{ablation}
\resizebox{0.65\linewidth}{!}{
\begin{tabular}{c c}
\hline
Method & Performance on GQA test-dev\\
\hline
w/o supervision & 51.31 \\
PAAN~\cite{paan} & 48.03 \\
HAN~\cite{han} & 49.96 \\
ASM~\cite{mining} & 52.96 \\
\hline
$L_{\Vec{\alpha}}$ + $L_{\Vec{r}}$ & 52.84 \\
AiR-M (w/o $L_{\Vec{\alpha}}$) & 50.01 \\
AiR-M (w/o $L_{\Vec{r}}$) & 50.33 \\
AiR-M & \textbf{53.46} \\
\hline
\end{tabular}
}
\end{center}
\end{table} 

As shown in Table \ref{ablation}, with the single-glimpse supervision of both the attention and reasoning operations (\ie~$L_{\Vec{\alpha}}$ + $L_{\Vec{r}}$), the task performance increases marginally from 51.31 to 52.96, which is comparable with the ASM methods~\cite{mining}. With independent progressive supervision on either the reasoning operation (AiR-M (w/o $L_{\Vec{\alpha}}$)) or the attention (AiR-M (w/o $L_{\Vec{r}}$)), however, the task performances decrease by about 1\%. In comparison, with the joint progressive supervision of attention and reasoning operations (\ie~AiR-M), the task performance is improved by more than 2\%. The performance change indicates that it is essential to jointly supervise attention and reasoning, so that the model can develop sufficient understanding of the complex interactions between them. 

\subsection{Ablation Study of Hyperparameters}\label{sec:supp_ablation_para}

The objective function of the proposed AiR-M attention supervision method consists of three loss terms:
\begin{equation}\label{att_loss}
    L = L_{ans} + \theta \sum\limits_{t} L_{\Vec{\alpha_t}} +  \phi \sum\limits_{t}  L_{\Vec{r}_t}
\end{equation}

\noindent where $L_{ans}$, $L_{\Vec{\alpha_t}}$ and $L_{\Vec{r}_t}$ are the loss terms on answer, attention and reasoning operations, respectively. The three loss terms are linearly combined with two hyperparameters, \textit{i.e.} $\theta$ and $\phi$. For $\theta$, we follow the dynamic hyperparameter proposed in \cite{mining}, and define it as $\theta = 0.5 (1+ \cos(\pi \cdot Iter/C))$, where $Iter$ is the current iteration and $C$ denotes the maximal number of training iterations ($C=300$k in our experiments). In terms of $\phi$, we conduct experiments on the balanced validation set of GQA with different $\phi$ values from 0.01 to 10. Table \ref{ablation_para} reports the results using UpDown~\cite{updown} as the baseline.

\begin{table}
\begin{center}
% \scriptsize
\caption{Experimental results of models trained with different settings of the hyperparameter (\ie~$\phi$ for objective related to operation prediction). All reported results are on the GQA \cite{gqa} balanced validation set. Bold numbers indicate the best performance.}
\label{ablation_para}
\resizebox{0.5\linewidth}{!}{
\begin{tabular}{c c}
\hline
$\phi$ & Performance on GQA Validation\\
\hline
0.01 & 62.02 \\
0.1 & 62.11 \\
% $\phi=0.25$ & 61.96 \\
0.5 & \textbf{62.57} \\
% $\phi=0.75$ & 61.92 \\
1 & 61.93 \\
10 & 61.39 \\
\hline
\end{tabular}
}
\end{center}
\end{table} 

According to the results, the proposed method is relatively robust against various settings of $\phi$, and we empirically find that $\phi=0.5$ provides the best validation accuracy. However, setting an much larger weight to the operation objective (\eg~$\phi=10$) tends to hamper the learning of attention and results in a considerable performance drop.

\subsection{Qualitative Results of AiR-M Attention Supervision}
\label{sec:supp_qual}
\begin{figure}
% \vspace{-2em}
\centering
\includegraphics[width=0.99\linewidth]{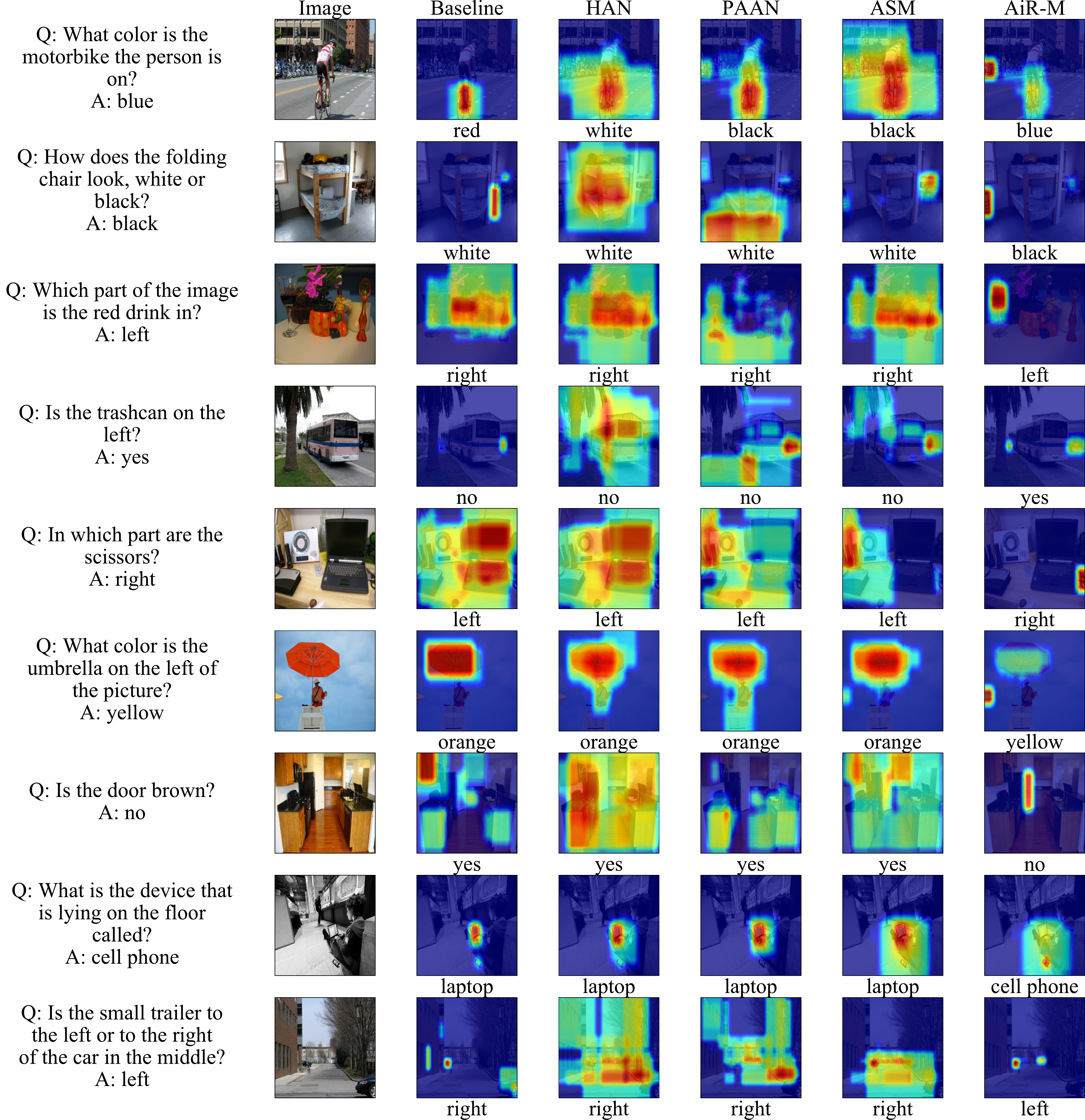}
\caption{Qualitative comparison between attention supervision methods, where Baseline
refers to UpDown~\cite{updown}. For each row, from left to right are the questions and the correct
answers, input images, and attention maps learned by different methods. The predicted
answers associated with the attentions are shown below their respective attention maps.}
\label{qual_supp}
\end{figure}

To further support our observations in the main paper (see \fig~6 of the main paper), we present additional qualitative examples of the AiR-M method, in comparison with the UpDown baseline and state-of-the-art attention supervision methods. As shown in \fig~\ref{qual_supp}, our method can accurately guide machine attention to focus on the ROIs related to the final answers (rows 1-7) as well as intermediate reasoning steps (rows 8-9).

\section{Supplementary Method}

\subsection{Decomposition of Reasoning Process}
\label{sec:supp_process}

As introduced in the main paper, we decompose the reasoning process into a sequence of atomic operations with regions of interest (ROIs) to conduct fine-grained evaluation of attentions for each step of the sequence. In this section, we describe this decomposition method in details:

\textbf{Deriving the Atomic Operations.} %As shown in Table 1 of the main paper, we define a set of atomic operations that are fundamental building blocks for visual reasoning. 
We derive the atomic operations (see Table 1 of the main paper) by characterizing and abstracting the complex operations in the functional programs of the GQA datset~\cite{gqa}. Specifically, we define each reasoning operation as a triplet, \textit{i.e.}, $<$operation, attribute, category$>$ and categorize the original operations in the program based on their semantic meanings: 
\begin{enumerate*}
    \item[(1)] For the original operations that exactly align with our definitions, we directly convert them into our triplet representation, for example, from ``filter size table'' to $<$filter, large/small, table$>$; 
    \item[(2)] If the original operations do not have an exact match, we convert them into our defined operations with similar semantic meanings. For example, we convert ``different color object A and object B'' to $<$compare, color, category A and category B$>$.
\end{enumerate*}
Most of the original GQA operations can be effectively converted into such triplet representations without loss of information. The triplets allow us to efficiently traverse the reasoning process by investigating the semantics of operations and their corresponding ROIs.

\textbf{Determining the ROIs of Each Operation.} The selection of ROIs depends on the semantics of the operations: 
\begin{itemize}
    \item \textbf{Select}: The ROIs belong to a specific category of objects. We query all objects in the scene graph and select those with the same category as defined in the triplet.
    \item \textbf{Query, Verify}: The ROIs are defined in a similar way as the ``select'' operation. The difference is that they are selected from the ROIs of the previous step, instead of the entire scene graph.
    \item \textbf{Filter}: The ROIs are a subset of the previous step's ROIs with the same attribute as defined in the triplet.
    \item \textbf{Compare, And, Or}: These operations are based on multiple groups of objects. Therefore, the ROIs are the combination of all the ROIs of the related previous reasoning steps.
    \item \textbf{Relate}: The ROIs are a combination of two groups of objects: the ROIs of the previous reasoning step and a specific category of objects from the scene graph.
\end{itemize}

% \textbf{Traversing the reasoning process.} With the triplets that encode the derived operations and the aforementioned scheme for selecting the ROIs, we sequentially traverse different steps in the reasoning process following their dependencies in the functional programs. Starting with the first step that preliminarily selects a group of objects (usually with a ``select'' operation), each operation will take the input objects from the outputs of dependent steps (or scene graph if the operation is ``select'' or ``relate'') and determine its ROIs. The proposed AiR-E score for each step is computed with the method discussed in the main paper, and the ROIs will be used as outputs passed on to the next operation. Following this procedure for every node, we eventually reach the final step that leads to the answer.

Some questions in GQA~\cite{gqa}, \eg~``Is there a red bottle on top of the table'' with answer ``no'', refer to non-existing objects. In such cases, we select the $k$ most frequently co-existent objects as the ROIs. Specifically, based on the GQA scene graphs, we first compute the frequency of co-existence between different object categories on the training set. Next, given a particular reasoning operation referring to an non-existing object, the top-$k$ ($k=20$) co-existing objects in the scene are selected as the corresponding ROIs.

\subsection{Attention Supervision Method (AiR-M)}
\label{sec:supp_supervision}

To demonstrate the effectiveness of the proposed AiR-M method for attention supervision, we apply it to three state-of-the-art VQA models, including UpDown \cite{updown} and MUTAN \cite{mutan} with standard visual attention, and BAN \cite{ban} with co-attention between vision and language. In this section, we present the implementation details of applying our method on different baseline models. 

\subsubsection{Application of AiR-M on UpDown~\cite{updown} and MUTAN~\cite{mutan}.} \label{visual}
\begin{figure}
\centering
\includegraphics[width=0.9\linewidth]{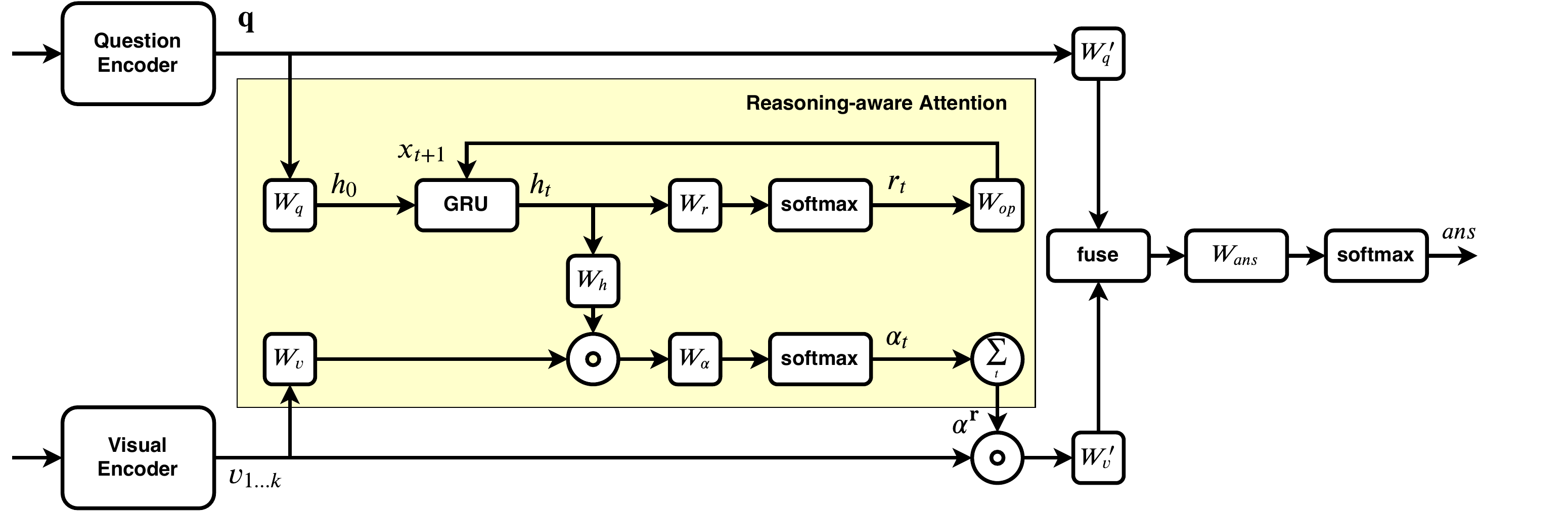}
\caption{High-level architecture of the proposed AiR-M method on models with standard visual attention.}
\label{fig:model}
\end{figure}

Many reasoning models, including UpDown and MUTAN, adopt the standard visual attention mechanism that computes attention weights for different units of visual input (\textit{i.e.} region proposals or spatial locations). %Since the AiR-M method aims at capturing various ROIs throughout the reasoning process, which aligns with the principal goal of visual attention, integrating our method is straightforward. 
To apply AiR-M on UpDown and MUTAN, we substitute their original attention mechanisms with the proposed one that jointly predicts the attention maps and the corresponding reasoning operations. Fig. \ref{fig:model} shows the high-level architecture of a model (\eg~UpDown\cite{updown} and MUTAN~\cite{mutan}) with its visual attention replaced with our proposed AiR-M.

Specifically, the AiR-M takes $q$ and $V$ as the inputs, and uses a Gated Recurrent Unit \cite{gru} (GRU) to sequentially predict the operations $\Vec{r}_t$ and the desired attention weights $\Vec{\alpha}_t$ at the $t$-th step. The attentions are aggregated into a final attention vector $\Vec{\alpha}^r$ to dynamically prioritize the visual features. 

At the beginning of the reasoning process, the hidden state of GRU $\Vec{h}_{0}$ with the question features $\Vec{q}$ is defined as:
\begin{equation}
    \Vec{h}_{0} = \Mat{W}_{q} \Vec{q},
\end{equation}
\noindent where $\Mat{W}_{q}$ represents trainable weights. We update the hidden state $\Vec{h}_{t}$, and simultaneously predict the reasoning operation $\Vec{r}_{t}$ and attention $\Vec{\alpha}_t$:
\begin{equation}
    \Vec{r}_{t} = \text{softmax} (\Mat{W}_{r} \Vec{h}_{t}),
\end{equation}
\begin{equation}\label{att_pred}
    \Vec{\alpha}_t = \text{softmax}( \Mat{W}_{\alpha} (\Mat{W}_{v} \Vec{v} \circ \Mat{W}_{h} \Vec{h}_{t}) )
\end{equation}
\noindent where $\Mat{W}_{r}, \Mat{W}_{\alpha}, \Mat{W}_{h}$ are all trainable weights, and $\circ$ is the Hadamard product. The next step input $\Vec{x}_{t+1}$ is computed with the predicted operation:
\begin{equation}
    \Vec{x}_{t+1} = \Mat{W}_{op} \Vec{r}_{t}
\end{equation}
\noindent where $\Mat{W}_{op}$ represents the weights of an embedding layer. 
By iterating over the whole sequence of reasoning steps, we compute the aggregated reasoning-aware attention
\begin{equation}
  \Vec{\alpha}^r =  \sum\limits_{t} \Vec{\alpha}_t / T
\end{equation}
that takes into account all the intermediate attention weights along the reasoning process, where $T$ is the total number of reasoning steps. With the supervision from the ROIs for different reasoning steps, the model is able to adaptively aggregate attention over time to perform complex visual reasoning.

Finally, following the original multi-modal fusion methods of UpDown~\cite{updown} and MUTAN~\cite{mutan}, we use $\Vec{\alpha}^r$ to determine the contributions of visual features via a dynamic weighting scheme, and the final answer is predicted based on the question and the attended visual features:
\begin{equation}
    ans  =  \text{softmax} \left(\Mat{W}_{ans} \ \text{fuse}(\Mat{W}_{v'} \sum\limits_{i} \Vec{\alpha}^r_{i} \Vec{v}_{i}, \Mat{W}_{q'}  \Vec{q}) \right)
\end{equation}
\noindent where $i$ denotes the index of visual features (\eg~region proposals or spatial locations), and the $\text{fuse}(\cdot)$ operator represents multi-modal fusion used in the baseline models (\eg~low-rank bilinear in UpDown~\cite{updown} and Tucker decomposition in MUTAN~\cite{mutan}).

\subsubsection{Application of AiR-M on BAN~\cite{ban}.}
To demonstrate the generality of our method, we further apply the AiR-M on a multi-glimpse co-attention model, \ie~BAN~\cite{ban}. Previous attention supervision methods~\cite{paan,han,mining} typically consider attention as a single-output mechanism, and have difficulty generalizing to multi-glimpse co-attention with multiple attention maps measuring the correlation between vision and language. Differently, our AiR-M method decomposes the reasoning process into a set of reasoning steps that naturally align with the multi-glimpse structure. Therefore, AiR-M with attention and reasoning supervision can guide the models to capture various ROIs with multi-glimpse attention. 

% Different from the visual attention that focuses on determining the contributions of visual features, co-attention bridges multiple modalities, \textit{e.g.}, vision and language, and correlates them by assigning weights for their pairwise relationship (\textit{e.g.}, a word to a spatial location). Moreover, many models with co-attention, including BAN \cite{ban}, make use of a multi-glimpse structure with multiple groups of attention weights. These characteristics make it difficult to directly apply previous attention supervision methods \cite{paan,han,mining}. Differently, our AiR-M method decomposes the reasoning process into a set of reasoning steps that naturally align with the multi-glimpse structure. Therefore, AiR-M with attention and reasoning supervision can guide the models to capture various ROIs with multi-glimpse attention.

Specifically, instead of using a fixed number of glimpses, we dynamically determine the number of glimpses based on the reasoning process. Following the same process as above for visual attention, we jointly compute the reasoning operation $\Vec{r}_{t}$ and corresponding attention $\Vec{\alpha}_t$ for reasoning step $t$. The attention $\Vec{\alpha}_t$ is applied to the visual features before computing the co-attention via $\Vec{v}'_t = \sum \Vec{\alpha}_t \Vec{v}$.

\subsubsection{Derivation of Ground-truth Attention Weights.}
The ground-truth attention weights used in the training objective for attention prediction (\ie~$L_{\Vec{\alpha_t}}$ in Equation~\ref{att_loss}) are derived from the GQA annotations. Specifically, we first extract the ROIs for each operation, and then compute the Intersection of Union (IoU) between each ROI and each input region proposal~\cite{updown}. The attention weight for each input region proposal is defined as the sum of its IoUs with all ROIs. Finally, the ground-truth attention weights of all input region proposals are normalized with their sum.

\subsubsection{Other Implementation Details.}
We train all of the models following the original settings proposed in the corresponding papers \cite{updown,mutan,ban}. Please refer to the original papers for further details. Since the original settings are designed for the VQA~\cite{vqa2} dataset, we make two modifications to accommodate the differences between GQA~\cite{gqa} and VQA~\cite{vqa2}: 
\begin{enumerate*}
    \item[(1)] We use batch size 150 for UpDown~\cite{updown} and MUTAN~\cite{mutan}, which tends to provide better results than the original settings for all models; 
    \item[(2)] Instead of using $G=8$ glimpses in BAN~\cite{ban} which leads to a severe overfitting, we follow \cite{gqa} (\ie~application of another multi-glimpse model MAC~\cite{mac}) and use $G=4$.
\end{enumerate*}

\end{document}